\documentclass{article}
\usepackage[margin=1in]{geometry}
\usepackage[T1]{fontenc}
\usepackage{lmodern}
\usepackage{microtype}
\usepackage{authblk}

\usepackage{natbib}
\usepackage{times}

\usepackage{amsmath,amsfonts,bm}

\def\eqref#1{equation~\ref{#1}}

\def\1{\bm{1}}

\DeclareMathAlphabet{\mathsfit}{\encodingdefault}{\sfdefault}{m}{sl}
\SetMathAlphabet{\mathsfit}{bold}{\encodingdefault}{\sfdefault}{bx}{n}

\usepackage{hyperref}
\usepackage{url}
\usepackage{booktabs}
\usepackage{graphicx}
\usepackage{xcolor}
\usepackage{enumitem}
\usepackage{subcaption}
\usepackage{amsmath}
\usepackage{multirow}
\usepackage{array}
\usepackage{algorithm}
\usepackage{algpseudocode}

\definecolor{colorGrad}{HTML}{2E86AB}
\definecolor{colorRand}{HTML}{27AE60}
\definecolor{colorWanda}{HTML}{C0392B}
\definecolor{colorNone}{HTML}{7F8C8D}

\title{Static Gradient Attribution Underperforms a\\
Density-Matched Random Mask on Loss-Based Forgetting
Within LoRA's B-Matrix}

\author[1]{Yu Sun\thanks{Corresponding author: yu.sun@alumni.upenn.edu}}
\author[2]{Junwei Zhou}
\author[3]{Zuodong Xiang}
\author[4]{Yike Zhang}
\author[5]{Pengcheng Xu}
\author[6]{Xinyu Guan}
\author[7]{Ruoyun Ma}
\author[8]{Hailu Xu}
\affil[1]{University of Pennsylvania}
\affil[2]{Matter Innovation Inc.}
\affil[3]{University of California, Davis}
\affil[4]{Johns Hopkins University}
\affil[5]{University of California, Irvine}
\affil[6]{University of Glasgow}
\affil[7]{Stanford University}
\affil[8]{California State University, Long Beach}
\date{}

\begin{document}

\maketitle

\begin{abstract}
Parameter-efficient continual learning updates a restricted subset of adapter parameters to incorporate sequential tasks while limiting catastrophic forgetting. Gradient attribution is a common rule for selecting that subset, but it requires an additional backward pass over calibration data. We identify a parameter-pool confound in comparisons between initialization-time gradient masks and unrestricted random masks under standard zero-initialized LoRA ($B=0$): the gradient with respect to $A$ is identically zero ($\partial\mathcal{L}/\partial A=0$), so gradient masks select strictly from the $B$ matrix whereas unrestricted random masks draw from both $A$ and $B$. Such a comparison changes both the selection rule and the candidate parameter pool. Isolating the selection criterion by restricting both approaches to the same parameter pool, we find a negative result: a uniform random mask consistently achieves lower loss-based backward transfer than gradient attribution across seven seeds on Qwen2.5-3B. The effect is concentrated on the medical task, which contributes 80\% of the aggregate; code and math move in the same direction, and legal shows no reliable difference. This advantage is consistent with a stability--plasticity trade-off under the tested task sequence, in which random masking slightly reduces single-task fitting. On the two discrete-option tasks (Medical and Legal), the mean accuracy-BWT difference is not statistically distinguishable from zero. The 95\% interval extends to $4.80$ percentage points in the gradient mask's favour, so the pre-specified 4-point non-inferiority criterion is not met. On loss-based forgetting, Random-B consistently outperforms the density-matched gradient mask across seven seeds; on accuracy, the evidence remains inconclusive at $n=7$. The attribution pass is therefore not justified by the loss-based forgetting evidence, while its accuracy consequences require a better-powered evaluation.

\end{abstract}

\section{Introduction}
\label{sec:intro}

Parameter-efficient adapters such as LoRA~\citep{hu2021lora, dettmers2023qlora} make it possible to incorporate a sequence of tasks while updating only a small fraction of a language model. Sequential updates can nevertheless overwrite capabilities acquired on earlier tasks, the classical problem of catastrophic forgetting~\citep{mccloskey1989catastrophic, french1999catastrophic}.
Continual learning for LoRA adapters updates only a chosen subset of parameters, and gradient attribution is the common rule for choosing it: a backward pass over held-out calibration data scores each parameter, and the highest-scoring ones are updated. Choosing the subset uniformly at random costs nothing instead, so whether the attribution pass earns its cost has a direct engineering answer.

Standard zero-initialized LoRA introduces a potential confound. With $B = 0$ at initialization, $\partial \mathcal{L}/\partial A = 0$, so gradient attribution assigns zero importance to every \texttt{lora\_A} parameter and a gradient mask selects only from \texttt{lora\_B}, while an unrestricted random baseline draws from both factors together. Any such comparison mixes the selection criterion with the parameter pool. Attribution-Guided CL~\citep{liu2026attributionguided} is the closest prior use of gradient attribution in LLM continual learning, but it estimates importance from a separately trained model; because $B\neq0$ at scoring time, the zero-initialization confound studied here does not apply to its attribution procedure.

We separate the two by restricting both arms to the same pool and parameter count. On Qwen2.5-3B with rank-16 LoRA over a five-task sequence we compare a uniformly random mask on \texttt{lora\_B} (Random-B) against a gradient mask renormalised to the same 10\% of \texttt{lora\_B} (Grad-B), so both train exactly the same parameters and differ only in how those parameters are chosen. Across seven seeds the random mask achieves the lower loss-based backward transfer, and the gap holds in every seed (Table~\ref{tab:d1}, Figure~\ref{fig:headline}). The aggregate gap is concentrated on the medical task, with code and math moving in the same direction and legal showing no reliable difference, a pattern consistent with a stability--plasticity trade-off: the random mask fits each task less well on its own diagonal yet retains earlier tasks better. Above all, this work makes two key contributions:

\begin{figure}[t]
\centering
\begin{minipage}[b]{0.85\textwidth}
\centering
\includegraphics[width=\textwidth]{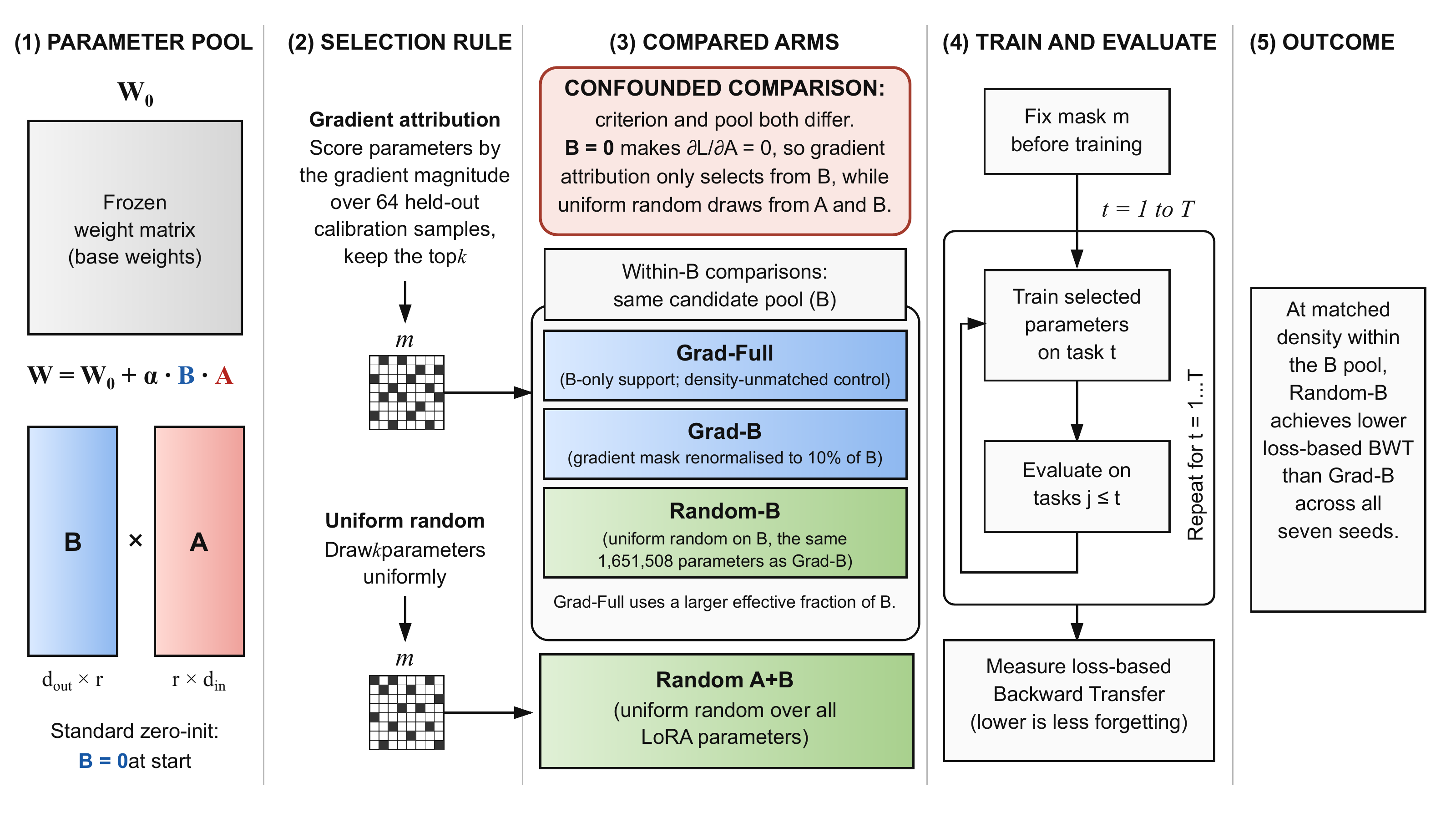}
\centerline{\footnotesize (a) Method and comparison.}
\end{minipage}\hfill
\begin{minipage}[b]{0.60\textwidth}
\centering
\includegraphics[width=\textwidth]{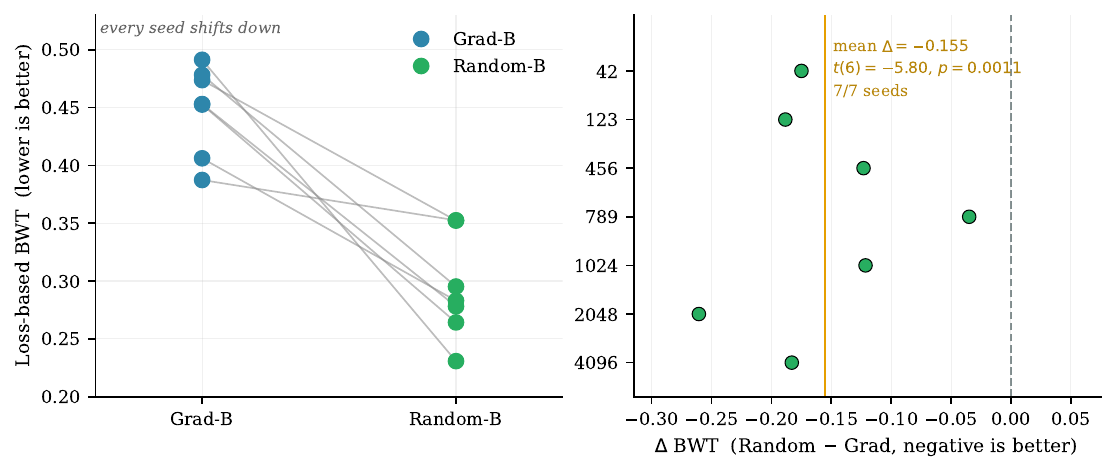}
\centerline{\footnotesize (b) Key effect (D1, seven seeds).}
\end{minipage}
\caption{The density-matched comparison. (a) Both arms draw 10\% of the rank-16 \texttt{lora\_B} pool, so they train exactly the same 1.6M parameters and differ only in how those coordinates are chosen; because $B=0$ at initialization, the gradient score is zero for every \texttt{lora\_A} coordinate, and any baseline that draws from both pools changes the pool as well as the criterion. (b) Under that matched density the random mask achieves lower loss-based BWT than the gradient mask in all seven seeds.}
\label{fig:headline}
\end{figure}

\textbf{A gradient-attribution confound and density-matched evaluation.}

We show that under standard zero-initialized LoRA, initialization-time gradient attribution assigns zero importance to \texttt{lora\_A}, so an unrestricted gradient-versus-random comparison changes both the selection rule and the candidate parameter pool. We resolve this confound with a within-\texttt{lora\_B}, density-matched design that holds the candidate pool and update budget fixed, leaving the selection criterion as the only varying factor.

\textbf{Practical validation and scope.}
We quantify the cost of the attribution pass and evaluate the same comparison using accuracy-based retention (accuracy-BWT). The accuracy-BWT interval spans zero and does not establish the pre-specified 4-point non-inferiority criterion, so the engineering recommendation is scoped to the loss-based evidence. Removing attribution eliminates a one-time 64-sample calibration forward--backward pass (64 steps), equal to 12.8\% of one 500-step task-training budget, together with the calibration-data requirement.

\section{Background and Related Work}
\label{sec:background}

\textbf{Continual learning and parameter importance.}
Continual learning aims to reduce catastrophic forgetting as models are updated on sequential tasks~\citep{mccloskey1989catastrophic,french1999catastrophic}. Importance-based methods such as EWC~\citep{kirkpatrick2017overcoming} and SI~\citep{zenke2017continual} protect parameters judged important to earlier tasks, while GPM~\citep{saha2021gpm}, OGD~\citep{farajtabar2020ogd}, and gradient-based selective-update methods~\citep{qiao2024pegp,wang2025gorp} constrain updates using gradient or parameter-importance structure. Pruning methods pose a related selection problem: Wanda~\citep{sun2024wanda} uses activation-aware saliency and SNIP~\citep{lee2019snip} scores connections before training. These lines motivate our question: once the candidate pool and update budget are controlled, does the \emph{selection rule} itself improve continual-learning retention.

\textbf{Random subspaces and intrinsic dimensionality.}
A complementary view is that the update budget may matter more than the exact coordinates selected. Random subnetworks can remain effective~\citep{frankle2019lottery,ramanujan2020whats}, fine-tuning often occupies a low-dimensional subspace~\citep{aghajanyan2020intrinsic}, and large fractions of delta parameters can be removed~\citep{yu2024dare}. Related PEFT results point in the same direction: HAT~\citep{serra2018hat} learns task-specific masks, \citet{su2024random} show that random masking can recover strong PEFT subnetworks, and \citet{biderman2024lora} find that LoRA forgets less than full fine-tuning. Because LoRA already constrains updates to a low-rank subspace, masking imposes a second restriction within it; our question is whether gradient attribution improves over a random restriction of the same size.

\paragraph{LoRA-specific continual learning and parameterization.}
LoRA-FA~\citep{zhang2023lorafa} freezes \texttt{lora\_A} and trains only \texttt{lora\_B}, whereas we freeze a random or gradient-selected subset of a standard LoRA. O-LoRA~\citep{wang2023olora} and InfLoRA~\citep{liang2024inflora} instead organize task updates into separated subspaces. PiSSA~\citep{meng2024pissa} is important to the scope of our analysis because it initializes $B\neq0$, so the zero-gradient confound we identify is specific to standard zero-initialized LoRA.

\paragraph{Closest and concurrent work.}
Most directly related, Attribution-Guided CL~\citep{liu2026attributionguided} uses gradient attribution in LLM continual learning, but estimates importance from a separately trained model and gates gradients continuously. Because $B\neq0$ at scoring time, our zero-initialization confound does not apply to that procedure; it nevertheless motivates the broader question of whether attribution improves retention relative to simpler controls. Concurrent results also point to substantial redundancy in adapter or sparse parameterizations: \citet{dieudonne2026adapter} study low-rank redundancy across task-specific LoRA adapters, while \citet{adewuyi2026multiple} find that random sparse subnetworks suffice for RLVR. Our contribution is complementary: we isolate the selection criterion itself by matching both the LoRA parameter pool and the update budget.

\paragraph{Adjacent applications of model adaptation and supervision.}
Beyond continual learning, efficient adaptation is relevant to resource-constrained retrieval-augmented language models, as discussed in a survey of energy-efficient on-device RAG with small language models~\citep{cheng2026toward}. Application-focused studies have also explored in-prompt process supervision for content moderation~\citep{liu-etal-2026-ips}, reasoning-enhanced multimodal domain adaptation~\citep{wang-etal-2025-reasoning-enhanced}, and audio-enhanced vision--language representation learning with latent-space broadening~\citep{10.1145/3711896.3737195}. These works concern different tasks and training settings; they provide broader context for adaptation and supervision, not prior evaluations of density-matched LoRA gradient masking.

\section{Methodology: Masking Mechanisms}
\label{sec:method}

\begin{table}[t]
\caption{Notation. Indices run over tasks $t = 1, \dots, T$ with $T = 5$, and over prior tasks $j \in \mathcal{P} = \{1, \dots, T-1\}$.}
\label{tab:notation}
\centering
\small
\begin{tabular}{@{}ll@{}}
\toprule
\textbf{Symbol} & \textbf{Meaning} \\
\midrule
$f_\theta$, $W_0$ & base model and its frozen pretrained weight matrix \\
$B$, $A$ & low-rank LoRA factors, $B \in \mathbb{R}^{d_{\text{out}} \times r}$, $A \in \mathbb{R}^{r \times d_{\text{in}}}$ \\
$r$, $\alpha$ & LoRA rank ($r = 16$) and scaling ($\alpha = 32$) \\
$W$ & adapted weight, $W = W_0 + \alpha BA$ \\
$t$, $T$ & task index and number of tasks ($T = 5$) \\
$\mathcal{D}_t$ & training split of task $t$ \\
$\mathcal{D}$ & held-out calibration set ($64$ Alpaca samples) \\
$m$ & binary mask over the trainable coordinates, $m \in \{0,1\}^{|A|+|B|}$ \\
$k$ & selection budget: number of coordinates with $m = 1$ \\
$g_i$ & gradient-attribution score of coordinate $i$ (\S\ref{sec:mask}) \\
$\mathcal{L}_j^{(\tau)}$ & evaluation loss on task $j$ after training through task $\tau$ \\
$\mathcal{P}$ & prior-task set $\{1, \dots, T-1\}$ \\
$\mathrm{BWT}$ & loss-based backward transfer, Eq.~\eqref{eq:bwt} \\
\bottomrule
\end{tabular}
\end{table}

\subsection{Problem Formulation}
\label{sec:problem}

We consider the standard sequential fine-tuning protocol for continual
learning. A base model $f_\theta$ is adapted
to task $t$ on its training split $\mathcal{D}_t$ and then evaluated on all
tasks $j \leq t$. Tasks arrive in a fixed order indexed by
$t = 1, \dots, T$, with $T=5$ throughout this paper.

We use LoRA adapters, which add a low-rank update
$\Delta W = BA$ to a frozen pretrained weight matrix $W_0$, where
$B \in \mathbb{R}^{d_{\mathrm{out}} \times r}$,
$A \in \mathbb{R}^{r \times d_{\mathrm{in}}}$, and
$r \ll \min(d_{\mathrm{out}}, d_{\mathrm{in}})$:
\begin{equation}
    W = W_0 + \alpha BA.
\end{equation}
Table~\ref{tab:notation} summarizes the notation used throughout.
A binary mask $m \in \{0,1\}^{|A|+|B|}$ restricts optimization to
coordinates for which $m_i=1$. The mask is constructed once from the
calibration set before the first task and remains fixed throughout the
task sequence (Algorithm~\ref{alg:mask}).

We compare two coordinate-selection rules: \emph{gradient attribution},
which ranks coordinates by their calibration-set gradient magnitude, and
\emph{uniform random selection}. For a fixed candidate parameter pool,
both methods activate the same budget $k$; they therefore differ only in
which coordinates are selected.

We measure forgetting using loss-based backward transfer:
\begin{equation}
\label{eq:bwt}
\mathrm{BWT}
=
\frac{1}{|\mathcal{P}|}
\sum_{j \in \mathcal{P}}
\left(
\mathcal{L}_j^{(T)}-\mathcal{L}_j^{(j)}
\right),
\end{equation}
where $\mathcal{P}=\{1,\dots,T-1\}$. This is the mean increase in
evaluation loss on prior tasks after completing the full sequence;
lower values indicate less forgetting.

\subsection{The $B=0$ Gradient-Attribution Confound}
\label{sec:b0}

Standard LoRA initializes $B=0$. For
\begin{equation}
    W = W_0 + \alpha BA,
\end{equation}
the gradient with respect to $A$ is
\begin{equation}
\label{eq:a_grad}
    \frac{\partial \mathcal{L}}{\partial A}
    =
    \alpha B^\top
    \frac{\partial \mathcal{L}}{\partial W}.
\end{equation}
Hence, at initialization,
\begin{equation}
    B=0
    \quad\Longrightarrow\quad
    \frac{\partial \mathcal{L}}{\partial A}=0.
\end{equation}
Gradient attribution therefore assigns zero importance to every
\texttt{lora\_A} coordinate and selects exclusively from
\texttt{lora\_B}. In contrast, an unrestricted random mask samples from
the full LoRA parameter pool, $\texttt{lora\_A}\cup\texttt{lora\_B}$.
A direct comparison between these masks consequently changes two factors
at once: the \emph{selection criterion} and the \emph{candidate parameter
pool}.

To isolate the selection criterion, the candidate pool must therefore be
matched. Restricting gradient attribution explicitly to
\texttt{lora\_B} is a null operation under zero initialization, whereas
restricting the random baseline to \texttt{lora\_B} removes the pool
mismatch. Our primary comparison accordingly evaluates gradient and
random selection within the same \texttt{lora\_B} pool; Section~\ref{sec:d1}
additionally matches their selection density.

This scope excludes Attribution-Guided CL~\citep{liu2026attributionguided},
which computes attribution from a separately trained model. Because
$B\neq0$ at scoring time in that procedure, the zero-initialization
confound identified here does not apply to it.

This confound is specific to standard zero-initialized LoRA. For example,
PiSSA~\citep{meng2024pissa} initializes both $A$ and $B$ from principal
singular values, so $B \neq 0$ and Equation~\ref{eq:a_grad} does not
force the gradient with respect to $A$ to vanish. LoRA+~\citep{hayou2024loraplus}
addresses related asymmetric optimization dynamics between the two LoRA
factors. Whether gradient attribution is beneficial under non-zero
initialization remains an open question.

\subsection{Density-Matched Mask Construction}
\label{sec:mask}

Let $\mathcal{S}$ denote the candidate set of LoRA coordinates from which
a mask is selected. For gradient attribution, each coordinate
$i \in \mathcal{S}$ is scored by its mean absolute gradient over the
held-out calibration set $\mathcal{D}$:
\begin{equation}
\label{eq:grad_score}
g_i
=
\mathbb{E}_{x \sim \mathcal{D}}
\left[
\left|
\frac{\partial \mathcal{L}(x)}{\partial w_i}
\right|
\right].
\end{equation}
Given a selection budget $k$, the binary mask activates the $k$
highest-scoring coordinates:
\begin{equation}
\label{eq:topk}
m_i
=
\mathbf{1}
\left[
i \in
\operatorname{TopK}
\left(
\{g_j\}_{j\in\mathcal{S}},\, k
\right)
\right].
\end{equation}

For uniform random selection, we instead draw independent scores
\begin{equation}
\label{eq:random_score}
u_i \sim \mathrm{Uniform}(0,1),
\qquad i \in \mathcal{S},
\end{equation}
and apply the same top-$k$ operator in Equation~\ref{eq:topk} with
$u_i$ in place of $g_i$. Thus, once the candidate pool
$\mathcal{S}$ and budget $k$ are fixed, the two methods differ only in
the rule used to rank coordinates.

\paragraph{Density-matched comparison.}
Following the pool confound established in \S\ref{sec:b0}, our primary
comparison restricts both selection rules to the
\texttt{lora\_B} coordinate pool, denoted by $\mathcal{B}$.
Both headline arms activate exactly 10\% of this pool:
\begin{equation}
\label{eq:d1_budget}
k_{\mathrm{D1}}
=
1{,}651{,}508.
\end{equation}
\emph{Grad-B} applies Equation~\ref{eq:grad_score} within
$\mathcal{B}$ and selects the top $k_{\mathrm{D1}}$ coordinates,
whereas \emph{Random-B} selects the same number of coordinates
uniformly at random from $\mathcal{B}$. The two arms therefore have
the same candidate pool and the same number of trainable coordinates;
the selection criterion is the only varying factor.

\paragraph{Reference and control arms.}
For comparison with the conventional design, \emph{Grad-Full} applies
gradient attribution with the full LoRA coordinate set
$\mathcal{A}\cup\mathcal{B}$ as its nominal candidate pool and uses a
10\% full-pool budget of 2,993,357 coordinates. Under standard
zero initialization, its selected support nevertheless lies entirely
within $\mathcal{B}$ by \S\ref{sec:b0}; because its budget is larger
than $k_{\mathrm{D1}}$, Grad-Full is pool-matched to Random-B in
effective support but not density-matched. \emph{Random A+B} draws the
same 10\% full-pool budget uniformly from
$\mathcal{A}\cup\mathcal{B}$ and therefore constitutes the conventional
pool-unmatched random baseline.

We additionally use two auxiliary controls where relevant.
\emph{No mask} is standard LoRA with all adapter coordinates updated.
\emph{Uniform-B} uses the same \texttt{lora\_B} budget as the matched
comparison while distributing selections uniformly across LoRA modules;
it is used only as a module-concentration control in
\S\ref{sec:module}.

Algorithm~\ref{alg:mask} summarizes mask construction and sequential
training.
\section{Experimental Setup}
\label{sec:setup}

\subsection{Models, Benchmarks \& Training Details}
\label{sec:models}

Qwen2.5-3B~\citep{qwen2024qwen25} in 4-bit NF4 (bitsandbytes), LoRA rank 16 ($\alpha$=32), AdamW (lr $2\times10^{-4}$, warmup 0.03), 1 epoch, 500 train samples per task, batch size 1, max length 512; calibration uses 64 Alpaca samples. All methods operate on 29.9M trainable LoRA parameters (0.96\% of 3.1B); the full-pool top-10\% selects $2{,}993{,}357$, falling entirely within \texttt{lora\_B} ($B=0$) and activating $\sim$20\% of B, against 10\% for Random-B. The comparison is matched on pool but not density; \S\ref{sec:d1} equalizes it.

Five domains are used in a fixed order: Code (sahil2801/CodeAlpaca-20k), Medical (GBaker/MedQA-USMLE-4-options), Legal (lex\_glue, casehold subset), Math (openai/gsm8k), Creative (ChaoticNeutrals/Creative\_Writing-ShareGPT, 6.6K instruction-response pairs). Calibration: tatsu-lab/alpaca (64 samples); 100 eval samples per task.

\subsection{Evaluation Metrics}
\label{sec:metrics}

We use Loss-based BWT in preference to accuracy-based BWT because loss is continuous and more sensitive to the small differences between masking strategies. 
All reported forgetting metrics are loss-based, reported as mean $\pm$ std over seven seeds (42, 123, 456, 789, 1024, 2048, 4096); part of the runs used a second GPU. The accuracy side of the same comparison, together with the option log-likelihood scorer and its scope, is reported separately in \S\ref{sec:accuracy_branch}.

\begin{figure}[!b]
\centering
\includegraphics[width=0.55\textwidth]{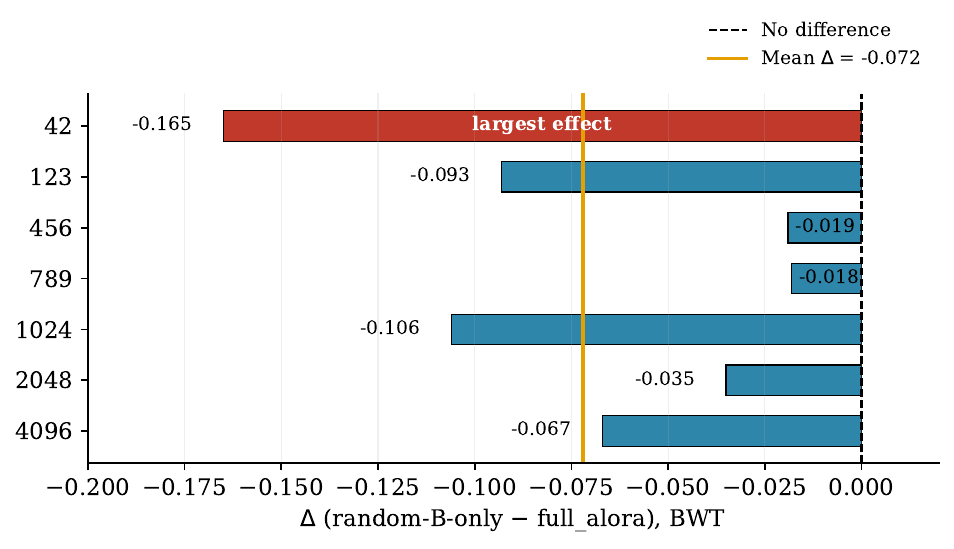}
\caption{Per-seed within-\texttt{lora\_B} differences ($\Delta = \text{Random-B} - \text{Grad-Full}$, loss-based BWT, 7 seeds). All seven are negative (Random-B forgets less), mean $\Delta = -0.072$ (orange line); seed 42 is the largest effect ($-0.165$). Leave-one-out excluding seed 42 remains significant (paired $t(5)$, $p = 0.015$).}
\label{fig:forest}
\end{figure}

\begin{table}[!t]
\caption{\textbf{Headline result (D1, density-matched).} Both arms select the same 1{,}651{,}508 parameters (10\% of \texttt{lora\_B}); only the criterion differs. Per-seed loss-based BWT and $\Delta = \text{Random-B} - \text{Grad-B}$ (negative = random-B forgets less), seven sessions, one seed each. The last column is the density-\emph{unmatched} contrast from the same sessions, that is, Random-B against the unrestricted gradient mask ($\Delta = \text{Random-B} - \text{Grad-Full}$) rather than against Grad-B.}
\label{tab:d1}
\centering
\small
\begin{tabular}{@{}lccrr@{}}
\toprule
\textbf{Seed} & \textbf{Random-B BWT $\downarrow$} & \textbf{Gradient-B BWT $\downarrow$} & \textbf{$\Delta$ (matched)} & \textbf{$\Delta$ (unmatched)} \\
\midrule
42  & $\mathbf{0.278}$ & 0.453 & $-0.175$ & $-0.220$ \\
123 & $\mathbf{0.264}$ & 0.453 & $-0.188$ & $-0.119$ \\
456 & $\mathbf{0.283}$ & 0.406 & $-0.123$ & $-0.131$ \\
789 & $\mathbf{0.352}$ & 0.387 & $-0.035$ & $-0.074$ \\
1024 & $\mathbf{0.352}$ & 0.474 & $-0.121$ & $-0.109$ \\
2048 & $\mathbf{0.231}$ & 0.491 & $-0.261$ & $-0.187$ \\
4096 & $\mathbf{0.295}$ & 0.478 & $-0.183$ & $-0.166$ \\
\midrule
Mean & $\mathbf{0.294}$ & 0.449 & $-0.155$ & $-0.144$ \\
\bottomrule
\end{tabular}
\end{table}

\section{Experiments and Main Results}
\label{sec:experiments}

\subsection{Confounded and Density-Matched Comparisons}
\label{sec:confounded}

A first comparison follows the usual design, in which each method draws from its own pool. Under that design gradient attribution shows no advantage: its loss-based BWT is $0.437 \pm 0.061$ against $0.428 \pm 0.019$ for a shared random mask over five seeds, a seed-dependent difference that is not significant (Table~\ref{tab:cl_3b}, Appendix~\ref{app:null_stats}). Both methods still reduce forgetting relative to an unmasked adapter, but the comparison is uninformative about the selection rule, for the reason set out in \S\ref{sec:b0}: gradient attribution selects only \texttt{lora\_B} while the random baseline draws from both pools, so the design varies pool and criterion together. The result is reported for completeness and is distinct from the seven-seed comparison that follows, the two coming from different sessions whose absolute levels are not comparable (Appendix~\ref{app:nf4_drift}).

Holding the pool fixed changes the picture. We introduce a Random-B control drawn from \texttt{lora\_B} alone, matching the gradient mask's pool, and compare it against a gradient mask restricted to the same pool. Each seed is an independent session, so absolute BWT values drift between runs (Appendix~\ref{app:nf4_drift}) and only within-session paired observations are used. This comparison is matched on pool but not on density, since the gradient mask activates nearly twice as many \texttt{lora\_B} coordinates as Random-B. Even so, the random mask achieves the lower loss-based BWT in all seven seeds, with a mean difference of $-0.072$ that survives dropping the largest-effect seed (two-sided paired $t(5)=-3.62$, $p = 0.015$; Figure~\ref{fig:forest}, Appendix~\ref{app:within_b}). This value is the direct recomputation on the six retained paired contrasts and corrects the $p=0.029$ value reported in an earlier draft. That the direction is unchanged when the pool is fixed is the first indication that the criterion, rather than the pool, is what the earlier comparison was measuring.

\subsection{Density-Matched Control}
\label{sec:d1}

Pool and criterion are separated cleanly only when density is matched as well. The comparison above leaves the gradient arm activating about 20\% of the pool against 10\% for Random-B, so the two arms still differ in how many coordinates they train. D1 removes that difference, with both arms drawing exactly 10\% of \texttt{lora\_B} and only the ranking rule differing. At matched density the random mask again achieves the lower loss-based BWT in every seed, by a larger margin than in the density-unmatched comparison (Table~\ref{tab:d1}). Under a common [Grad-B $-$ Random-B] convention, the mean contrast is $+0.155$ on the V100 stack and $+0.152$ on the A800 stack. Thus the contrast reproduces at a similar magnitude in a second hardware/software environment; in a separate three-seed sensitivity check, reversing the execution order of the two arms does not reverse the Random-B--Grad-B contrast (Appendix~\ref{app:nf4_drift}).

\subsection{Task-Level Decomposition}
\label{sec:decomposition}

\begin{figure}[!htbp]
\centering
\includegraphics[width=0.9\textwidth]{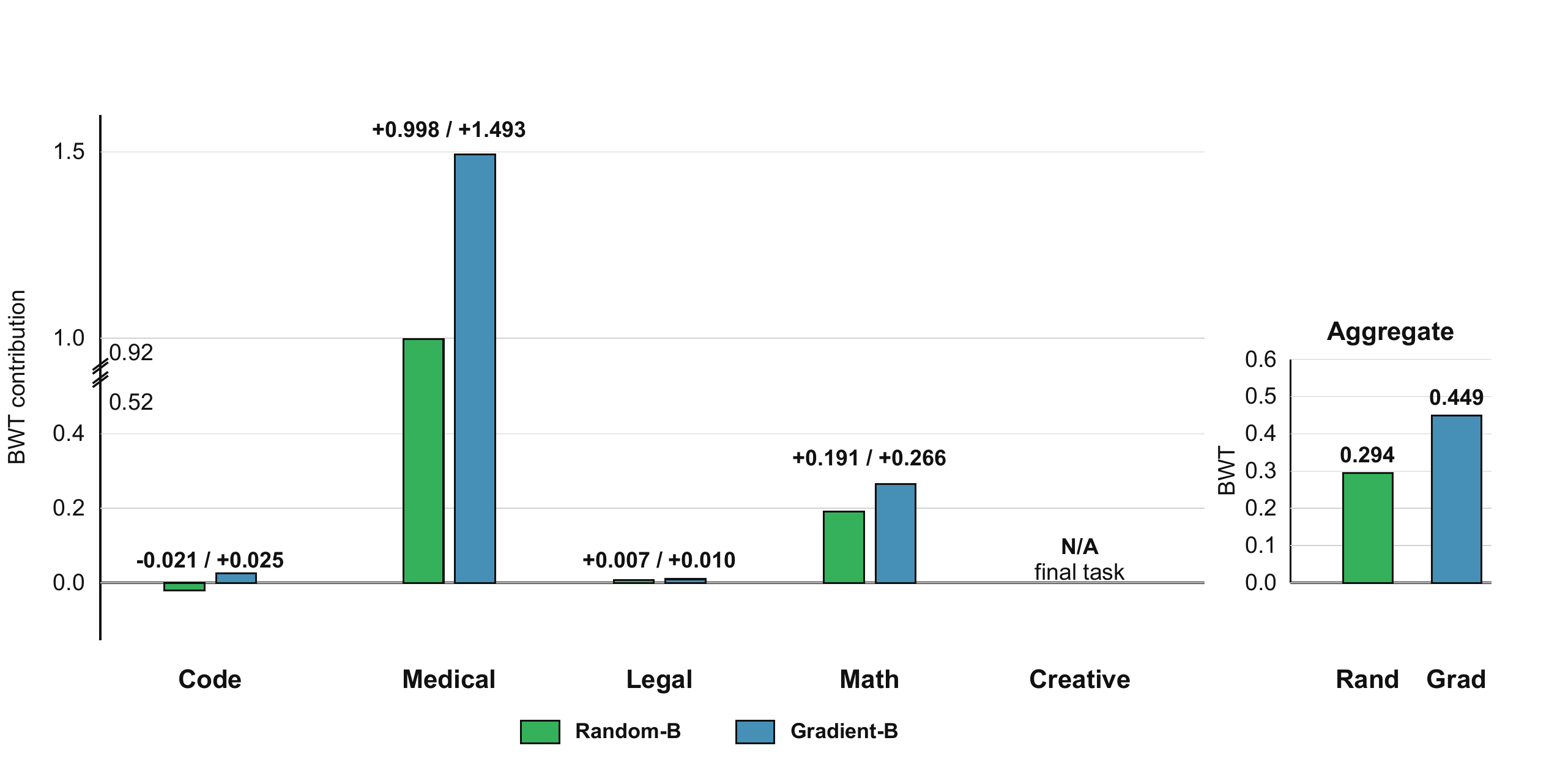}
\caption{The density-matched D1 per-task decomposition, averaged over seven seeds.
For each prior task, the bars show the loss-based BWT contribution of
Random-B and Grad-B, with the corresponding values
reported above each pair. 
Because the task-level contributions span roughly two orders of magnitude, the left panel uses a broken $y$-axis: the slashes mark the omitted interval from 0.52 to 0.92, and the labels above the Medical bars report their full values ($+0.998$ and $+1.493$) rather than truncated heights.
Creative is the final task and therefore has no backward-transfer contribution. The right panel shows
the aggregate loss-based BWT on a separate scale. Random-B exhibits less loss-based
forgetting on Code, Medical, and Math; the Legal difference is negligible and changes
sign across seeds. Most of the aggregate difference is concentrated on Medical.
Per-seed results are reported in Appendix~\ref{app:d1_per_task}.
}
\label{fig:per_task}
\end{figure}

The aggregate gain is not spread evenly across the five tasks, and where it concentrates bears on how the effect should be read. Medical carries about four fifths of it, while code and math contribute the remainder in the same direction; on legal the arms are indistinguishable. 
Two explanations can be assessed from the same data. First, the gradient arm's loss on each task's own diagonal is lower than the random arm's in every seed and task, yet it subsequently forgets more, so the two arms differ in retention rather than initial learning capacity. The same seven within-session pairs also show lower mean final held-out loss over the four prior tasks for Random-B: the Grad-B minus Random-B endpoint difference is $+0.0813$ (paired $t(6)=3.03$, $p=0.023$; 6/7 seeds). This endpoint result is less robust to leave-one-out deletion than the headline BWT contrast, but it shows that the effect is not merely a BWT normalization artifact (Appendix~\ref{app:d1_diagonal}). By contrast, restoring the gradient arm's full budget produces seed-dependent results: at seed 42 the density-matched gradient arm outperforms the unrestricted one, whereas the ordering reverses at the seven-seed mean. It therefore does not provide a stable explanation of the headline effect.

\subsection{Accuracy-Based Validation}
\label{sec:accuracy_branch}

\begin{figure}[!htbp]
\centering
\includegraphics[width=0.6\textwidth]{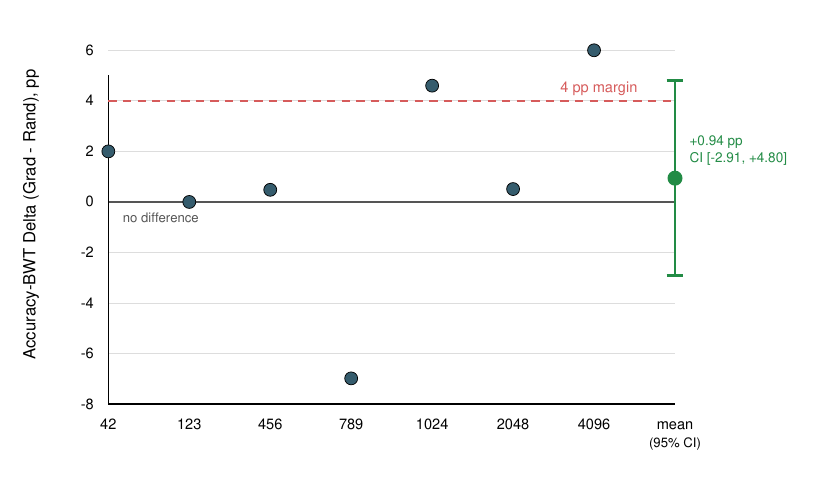}
\caption{Accuracy-based validation. Blue points are the seven per-seed accuracy-BWT differences (Grad-B minus Random-B, percentage points) on the two discrete-option tasks, spanning $-6.98$ to $+6.00$ pp; green is the seven-seed mean with its 95\% interval $[-2.91,+4.80]$ pp. The plotted quantity is a difference of accuracy-BWT (mean final-minus-learned accuracy over prior tasks), a retention measure, not a difference of final accuracies. The interval spans zero, so no statistically reliable accuracy difference is detected, and it reaches past the 4 pp non-inferiority margin, so that criterion is not met. Per-seed values are in Appendix~\ref{app:accuracy_seeds}.}
\label{fig:accuracy_ci}
\end{figure}

Loss-based BWT is a proxy for forgetting rather than the quantity of ultimate interest, so we evaluate
the same masking comparison on task accuracy and ask whether the accuracy ordering agrees with
the loss ordering. The evaluation reuses the primary protocol and the corrected task evaluators over
the same seven seeds and 100 evaluation items per task, retraining from the base model on an A800
rather than re-scoring a checkpoint. Accuracy is scored only for the two tasks with discrete options
(Medical, MedQA four options; Legal, CaseHOLD), so every quantity below concerns those two.

Across seven seeds, the accuracy-BWT difference averages $+0.94$ percentage points with a
95\% interval of $[-2.91,+4.80]$ pp that spans zero. Here accuracy-BWT is the mean
final-minus-learned accuracy over the prior discrete-option tasks, so it measures retention on the
same footing as loss-based BWT and is not a difference of final accuracies. The individual seeds
range from $-6.98$ to $+6.00$ pp, and it is that spread rather than a large mean that makes the
interval wide at $n=7$. A 4 pp non-inferiority margin was specified in advance, and the measured
upper bound exceeds it, so non-inferiority is not established.

On the loss side, the same masks are separated: Grad-B forgets more in all seven seeds by a mean
of $+0.152$ on the A800 loss runs. This agrees in mean to within $0.003$ with the V100 estimate
after aligning sign conventions ($+0.155$ under [Grad $-$ Random]). Loss therefore resolves a
difference the accuracy measurement does not; the accuracy interval bounds that difference rather
than contradicting it.

Per-seed values and the interval construction are reported in Appendix~\ref{app:accuracy_seeds}.
Math is excluded from the accuracy read-out because its generation-based score is confounded by
truncation (Appendix~\ref{app:math_audit}).

\section{Analysis and Extensions}
\label{sec:analysis}

\subsection{Sparsity, Scale, and Cross-Architecture Generalization}
\label{sec:boundaries}

The headline comparison fixes the sparsity budget at 10\%. Sweeping it from 5\% to 30\% at a single seed, no tested level shows gradient attribution outperforming a shared random mask (Figure~\ref{fig:sparsity}, Appendix~\ref{app:sparsity}). The sweep is preliminary and inherits the pool confound rather than isolating the budget, so we report it as a boundary check on the direction of the effect rather than as evidence for the sparsity-constraint view.

No single-seed probe reverses the conclusion, but a reversed task order shifts BWT by more than any method difference, which is why every conclusion in this paper is scoped to the tested sequence (Table~\ref{tab:scale_obs}, Appendices~\ref{app:sparsity}--\ref{app:task_order}). On Phi-2, an architecturally distinct family, we additionally run a clean density-matched D1 replication: both arms select exactly 10\% of the same \texttt{lora\_B} pool and therefore differ only in the ranking rule. Random-B achieves lower loss-based BWT in all four seeds, with mean $\Delta=\text{Random-B}-\text{Grad-B}=-0.1152$ (SD $0.0173$), compared with $-0.1552$ on Qwen2.5-3B. The task-level decomposition also reproduces the primary pattern: Medical contributes 67.2\% of the Phi-2 aggregate contrast and is the largest contributor in every seed, versus 79.9\% on Qwen2.5-3B. Because the replication has only four seeds, we treat its parametric test as supporting evidence and base the cross-architecture claim primarily on the matched design and 4/4 directional replication; full per-seed results and the sign-test caveat are in Appendix~\ref{app:phi2_d1}.

\subsection{Module Concentration and Calibration Robustness}
\label{sec:module}

The gradient mask does not select uniformly across module types: within \texttt{lora\_B} it concentrates on output projections and to a lesser degree on \texttt{up\_proj}, together accounting for a little over half of all selections, while down projections receive well below the uniform rate. The pattern is consistent across calibration sources and peaks at middle layers (Appendix~\ref{app:mask_concentration}). That concentration invites the reading that it causes the gradient mask's higher BWT, and the \texttt{uniform\_B} controller tests it by spreading the same budget across modules. The controller contradicts the reading: three of four seeds land on the gradient side rather than the random side. Because those offsets are taken against different session baselines (Appendix~\ref{app:nf4_drift}), we treat the ablation as failing to establish the mechanism rather than as evidence against the concentration itself, and neither the within-\texttt{lora\_B} finding nor its density-matched control depends on it (Figure~\ref{fig:oproj_ablation_main}). The result is also insensitive to the calibration domain: pooling in-domain rather than out-of-domain calibration changes 42\% of the selected parameters, yet the loss-based BWT moves by less than a tenth of the between-seed spread (Appendix~\ref{app:calibration}).

\section{Conclusion}
\label{sec:conclusion}

At equal parameter count within the \texttt{lora\_B} pool, the random mask achieves lower loss-based BWT in every seed, an advantage that reproduces in mean and direction in a second hardware/software environment and is not an artifact of density. The same paired runs also show lower mean final held-out loss on the prior tasks for Random-B (Grad-B minus Random-B $=+0.0813$, $p=0.023$), although this endpoint result is less leave-one-out robust than the BWT contrast. On the loss-based evidence, the attribution pass is not justified: removing it eliminates a one-time 64-sample calibration forward--backward pass (64 steps), equal to 12.8\% of one 500-step task-training budget, and removes the calibration-data requirement. Accuracy-BWT remains inconclusive at seven seeds, with the pre-specified non-inferiority criterion unmet. All conclusions are scoped to the tested task order and static masks; a per-task adaptive mask and a better-powered accuracy evaluation are the most informative next tests.

\paragraph{Limitations.}
\label{sec:limitations}

The within-\texttt{lora\_B} finding is limited to static masks,
a single primary task order, and rank-16 LoRA under standard
zero initialization. Our primary conclusions are based on loss-based
BWT, with accuracy validation limited to the two discrete-option tasks.
Absolute BWT also varies across sessions, so all statistical claims rely
on within-session paired comparisons (Appendix~\ref{app:nf4_drift}).

\section*{Reproducibility Statement}

Experiments use Python 3.12, PyTorch 2.8.0+cu128 (4-bit NF4 with bf16 compute), Transformers 5.16.1, PEFT 0.20.0, bitsandbytes 0.50.2, datasets 5.0.1, and CUDA 12.8. The primary seven-seed within-\texttt{lora\_B} comparison, scale runs, and ablations use a single NVIDIA A800 80GB PCIe GPU. The D1 density-matched control and the M2 uniform-B control ran in a separate session on a Tesla V100-PCIE-32GB under an earlier environment (Python 3.12.3, PyTorch 2.3.0+cu121, Transformers 4.46.3, PEFT 0.13.2, bitsandbytes 0.43.3); as both are within-run paired comparisons, the hardware difference does not affect their paired results. Primary results use Qwen2.5-3B (seven seeds); cross-architecture within-B results use Phi-2 2.78B (four seeds); the confounded Phi-2 comparison uses five seeds under a matched 100-eval-sample protocol (an additional seed-42 observation used 20 eval samples and is reported separately); scale observations use Qwen2.5-0.5B/1.5B/7B (single seed). LoRA is applied to all linear projection layers.

Model training uses AdamW ($\beta_1{=}0.9$, $\beta_2{=}0.999$, weight decay 0.01) at learning rate $2\times10^{-4}$ with a linear warmup of 0.03 and cosine annealing over the per-task step budget; LoRA dropout 0.05, rank 16, $\alpha$=32, batch size 1, max length 512, one epoch per task, and top-$k$ 0.1 masking. Five sequential tasks are used: Code (code\_alpaca\_20k), Medical (medqa, 4-option MC), Legal (lexglue/case\_hold), Math (gsm8k), and Creative Writing (Creative\_Writing-ShareGPT). Each task uses 500 training and 100 evaluation samples; gradient masks use 64 out-of-domain Alpaca calibration samples. Seeds are 42, 123, 456, 789, 1024, 2048, and 4096; approximately 60 GPU-hours were used in total.

Boundary analyses (rank sweep, task order reversal, scale observations) are single-seed; a method-\emph{order} check over three seeds is reported in Appendix~\ref{app:nf4_drift}. An isolated-process protocol equivalence check confirms that running each method in a separate process reproduces the multi-method BWT to within $9.6 \times 10^{-5}$ (10$\times$ margin below the pre-specified $10^{-3}$ threshold). All reported forgetting metrics are loss-based BWT. Code and configuration files are released as anonymized supplementary material (\texttt{code\_anonymous.zip}): training and attribution scripts, five-task data preparation, mask generation, the analysis scripts computing BWT and the paired statistics, and pinned dependency manifests for both environments.

\section*{Ethics Statement}

This work studies parameter masking for continual learning and does not involve human subjects or sensitive data. It does not introduce new capabilities or datasets that raise ethical concerns. On data sources and licensing: the Code task uses CodeAlpaca-20k (CC BY 4.0), the Math task uses GSM8K (MIT), and the Medical and Legal tasks use MedQA and CaseHOLD, which are distributed on Hugging Face with no affirmative license declaration, so we use them for research evaluation only. The Creative task uses a synthetic creative-writing set composed of model-generated outputs rather than user conversations. The Alpaca data used to calibrate the gradient masks is released under CC BY-NC 4.0, a non-commercial license; we therefore make no commercial-use claim for models trained with it. The finding that random selection outperforms gradient attribution within LoRA's B-matrix may reduce computational cost in continual-learning deployments by removing the need for gradient-based parameter attribution. We do not foresee negative societal consequences specific to this contribution.

\section*{AI Use Disclosure}

AI coding assistants were used for experiment-script development and debugging, and AI assistance aided literature search and manuscript drafting and revision. The research design, experimental methodology, statistical analysis, and scientific conclusions are the authors' own work. All reported results were produced by the described experimental pipeline without AI-generated synthetic data; all analyses and claims were verified by the authors.

\bibliography{references}

@article{kirkpatrick2017overcoming,
  title = {Overcoming Catastrophic Forgetting in Neural Networks},
  author = {James Kirkpatrick and Razvan Pascanu and Neil Rabinowitz and Joel Veness and Guillaume Desjardins and Andrei A. Rusu and Kieran Milan and John Quan and Tiago Ramalho and Agnieszka Grabska-Barwinska and Demis Hassabis and Claudia Clopath and Dharshan Kumaran and Raia Hadsell},
  year = {2017},
  journal = {Proceedings of the National Academy of Sciences},
  volume = {114},
  number = {13},
  pages = {3521--3526},
  doi = {10.1073/pnas.1611835114},
  url = {https://arxiv.org/abs/1612.00796},
}

@inproceedings{zenke2017continual,
  title = {Continual Learning Through Synaptic Intelligence},
  author = {Zenke, Friedemann and Poole, Ben and Ganguli, Surya},
  year = {2017},
  booktitle = {Proceedings of the 34th International Conference on Machine Learning},
  series = {Proceedings of Machine Learning Research},
  volume = {70},
  pages = {3987--3995},
  url = {https://proceedings.mlr.press/v70/zenke17a.html},
}

@inproceedings{sun2024wanda,
  title = {A Simple and Effective Pruning Approach for Large Language Models},
  author = {Sun, Mingjie and Liu, Zhuang and Bair, Anna and Kolter, J. Zico},
  year = {2024},
  booktitle = {International Conference on Learning Representations},
  url = {https://openreview.net/forum?id=PxoFut3dWW},
}

@article{hu2021lora,
  title = {{LoRA}: Low-Rank Adaptation of Large Language Models},
  author = {Hu, Edward J. and Shen, Yelong and Wallis, Phillip and Allen-Zhu, Zeyuan and Li, Yuanzhi and Wang, Shean and Wang, Lu and Chen, Weizhu},
  year = {2021},
  journal = {arXiv preprint arXiv:2106.09685},
  url = {https://arxiv.org/abs/2106.09685},
}

@article{dettmers2023qlora,
  title = {{QLoRA}: Efficient Finetuning of Quantized {LLM}s},
  author = {Tim Dettmers and Artidoro Pagnoni and Ari Holtzman and Luke Zettlemoyer},
  year = {2023},
  journal = {arXiv preprint arXiv:2305.14314},
  url = {https://arxiv.org/abs/2305.14314},
}

@inproceedings{frankle2019lottery,
  title = {The Lottery Ticket Hypothesis: Finding Sparse, Trainable Neural Networks},
  author = {Frankle, Jonathan and Carbin, Michael},
  year = {2019},
  booktitle = {International Conference on Learning Representations},
  url = {https://arxiv.org/abs/1803.03635},
}

@inproceedings{meng2024pissa,
  title = {{PiSSA}: Principal Singular Values and Singular Vectors Adaptation of Large Language Models},
  author = {Meng, Fanxu and Wang, Zhaohui and Zhang, Muhan},
  year = {2024},
  booktitle = {Advances in Neural Information Processing Systems},
  volume = {37},
  pages = {121038--121072},
  url = {https://arxiv.org/abs/2404.02948},
}

@inproceedings{ramanujan2020whats,
  title = {What's Hidden in a Randomly Weighted Neural Network?},
  author = {Ramanujan, Vivek and Wortsman, Mitchell and Kembhavi, Aniruddha and Farhadi, Ali and Rastegari, Mohammad},
  year = {2020},
  booktitle = {2020 IEEE/CVF Conference on Computer Vision and Pattern Recognition},
  url = {https://openaccess.thecvf.com/content_CVPR_2020/html/Ramanujan_Whats_Hidden_in_a_Randomly_Weighted_Neural_Network_CVPR_2020_paper.html},
}

@article{french1999catastrophic,
  title = {Catastrophic Forgetting in Connectionist Networks},
  author = {Robert M. French},
  year = {1999},
  journal = {Trends in Cognitive Sciences},
  volume = {3},
  number = {4},
  pages = {128--135},
}

@incollection{mccloskey1989catastrophic,
  title = {Catastrophic Interference in Connectionist Networks: The Sequential Learning Problem},
  author = {McCloskey, Michael and Cohen, Neal J.},
  year = {1989},
  booktitle = {Psychology of Learning and Motivation},
  publisher = {Academic Press},
}

@inproceedings{aljundi2018memory,
  title = {Memory Aware Synapses: Learning what (not) to forget},
  author = {Aljundi, Rahaf and Babiloni, Francesca and Elhoseiny, Mohamed and Rohrbach, Marcus and Tuytelaars, Tinne},
  year = {2018},
  booktitle = {Proceedings of the European Conference on Computer Vision},
  pages = {139--154},
  url = {https://openaccess.thecvf.com/content_ECCV_2018/html/Rahaf_Aljundi_Memory_Aware_Synapses_ECCV_2018_paper.html},
}

@inproceedings{aghajanyan2020intrinsic,
  title = {Intrinsic Dimensionality Explains the Effectiveness of Language Model Fine-Tuning},
  author = {Armen Aghajanyan and Sonal Gupta and Luke Zettlemoyer},
  year = {2021},
  booktitle = {Proceedings of the 59th Annual Meeting of the Association for Computational Linguistics and the 11th International Joint Conference on Natural Language Processing},
  url = {https://arxiv.org/abs/2012.13255},
}

@article{qwen2024qwen25,
  title = {{Qwen2.5} Technical Report},
  author = {Qwen Team},
  year = {2024},
  journal = {arXiv preprint arXiv:2412.15115},
  url = {https://arxiv.org/abs/2412.15115},
}

@inproceedings{su2024random,
  title = {Random Masking Finds Winning Tickets for Parameter Efficient Fine-tuning},
  author = {Xu, Jing and Zhang, Jingzhao},
  year = {2024},
  booktitle = {Proceedings of the 41st International Conference on Machine Learning},
  url = {https://arxiv.org/abs/2405.02596},
}

@article{biderman2024lora,
  title = {{LoRA} Learns Less and Forgets Less},
  author = {Biderman, Dan and Portes, Jacob and Gonzalez Ortiz, Jose Javier and Paul, Mansheej and Greengard, Philip and Jennings, Connor and King, Daniel and Havens, Sam and Chiley, Vitaliy and Frankle, Jonathan and Blakeney, Cody and Cunningham, John P.},
  year = {2024},
  journal = {Transactions on Machine Learning Research},
  url = {https://arxiv.org/abs/2405.09673},
}

@inproceedings{wang2023olora,
  title = {{O-LoRA}: Orthogonal Subspace Learning for Language Model Continual Learning},
  author = {Wang, Xiao and Chen, Tianze and Ge, Qiming and Xia, Han and Bao, Rong and Zheng, Rui and Zhang, Qi and Gui, Tao and Huang, Xuanjing},
  year = {2023},
  booktitle = {Findings of the Association for Computational Linguistics: EMNLP 2023},
  pages = {10658--10671},
  url = {https://arxiv.org/abs/2310.14152},
}

@inproceedings{wang2025gorp,
  title = {Continual Gradient Low-Rank Projection Fine-Tuning for {LLM}s},
  author = {Wang, Chenxu and Lyu, Yilin and Sun, Zicheng and Jing, Liping},
  year = {2025},
  booktitle = {Proceedings of the 63rd Annual Meeting of the Association for Computational Linguistics},
  url = {https://aclanthology.org/2025.acl-long.721/},
}

@inproceedings{farajtabar2020ogd,
  title = {Orthogonal Gradient Descent for Continual Learning},
  author = {Farajtabar, Mehrdad and Azizan, Navid and Mott, Alex and Li, Ang},
  year = {2020},
  booktitle = {Proceedings of the 23rd International Conference on Artificial Intelligence and Statistics},
  series = {Proceedings of Machine Learning Research},
  volume = {108},
  pages = {3762--3773},
  url = {https://arxiv.org/abs/1910.07104},
}

@inproceedings{saha2021gpm,
  title = {Gradient Projection Memory for Continual Learning},
  author = {Saha, Gobinda and Garg, Isha and Roy, Kaushik},
  year = {2021},
  booktitle = {International Conference on Learning Representations},
  url = {https://arxiv.org/abs/2103.09762},
}

@inproceedings{yu2024dare,
  title = {Language Models are {Super Mario}: Absorbing Abilities from Homologous Models as a Free Lunch},
  author = {Yu, Le and Yu, Bowen and Yu, Haiyang and Huang, Fei and Li, Yongbin},
  year = {2024},
  booktitle = {Proceedings of the 41st International Conference on Machine Learning},
  url = {https://arxiv.org/abs/2311.03099},
}

@inproceedings{liang2024inflora,
  title = {{InfLoRA}: Interference-Free Low-Rank Adaptation for Continual Learning},
  author = {Liang, Yan-Shuo and Li, Wu-Jun},
  year = {2024},
  booktitle = {Proceedings of the IEEE/CVF Conference on Computer Vision and Pattern Recognition},
  pages = {23638--23647},
  url = {https://arxiv.org/abs/2404.00228},
}

@article{dieudonne2026adapter,
  title = {When One Adapter Speaks for Many: Discovering Low-Rank Redundancy in Continual Fine-Tuning},
  author = {Dieudonn{\'e}, Tanguy and Lanzillotta, Giulia and Simsar, Enis and Barinka, Louis and Hofmann, Thomas},
  journal = {arXiv preprint arXiv:2606.28117},
  year = {2026},
}

@article{adewuyi2026multiple,
  title = {The Multiple Ticket Hypothesis: Random Sparse Subnetworks Suffice for {RLVR}},
  author = {Adewuyi, Israel and Okibe, Solomon and Ivanov, Vladmir},
  journal = {arXiv preprint arXiv:2602.01599},
  year = {2026},
}

@article{liu2026attributionguided,
  title = {Attribution-Guided Continual Learning for Large Language Models},
  author = {Liu, Yazheng and Wan, Yuxuan and Xu, Rui and Zhang, Xi and Xie, Sihong and Xiong, Hui},
  journal = {arXiv preprint arXiv:2605.05285},
  year = {2026},
}

@article{hayou2024loraplus,
  title = {{LoRA+}: Efficient Low Rank Adaptation of Large Models},
  author = {Hayou, Soufiane and Ghosh, Nikhil and Yu, Bin},
  year = {2024},
  journal = {arXiv preprint arXiv:2402.12354},
  url = {https://arxiv.org/abs/2402.12354},
}

@inproceedings{serra2018hat,
  title = {Overcoming Catastrophic Forgetting with Hard Attention to the Task},
  author = {Serr\`{a}, Joan and Sur\'{i}s, D\'{i}dac and Miron, Marius and Karatzoglou, Alexandros},
  year = {2018},
  booktitle = {Proceedings of the 35th International Conference on Machine Learning},
  series = {Proceedings of Machine Learning Research},
  volume = {80},
  pages = {4548--4557},
  url = {https://arxiv.org/abs/1801.01423},
}

@inproceedings{lee2019snip,
  title = {{SNIP}: Single-Shot Network Pruning Based on Connection Sensitivity},
  author = {Lee, Namhoon and Ajanthan, Thalaiyasingam and Torr, Philip H. S.},
  year = {2019},
  booktitle = {International Conference on Learning Representations},
  url = {https://arxiv.org/abs/1810.02340},
}

@inproceedings{wang2020grasp,
  title = {Picking Winning Tickets Before Training by Preserving Gradient Flow},
  author = {Wang, Chaoqi and Zhang, Guodong and Grosse, Roger},
  year = {2020},
  url = {https://arxiv.org/abs/2002.07376},
}

@article{zhang2023lorafa,
  title = {{LoRA-FA}: Efficient and Effective Low Rank Representation Fine-tuning},
  author = {Zhang, Longteng and Zhang, Lin and Shi, Shaohuai and Chu, Xiaowen and Li, Bo},
  year = {2023},
  journal = {arXiv preprint arXiv:2308.03303},
  url = {https://arxiv.org/abs/2308.03303},
}

@article{qiao2024pegp,
  title = {Gradient Projection For Continual Parameter-Efficient Tuning},
  author = {Qiao, Jingyang and Zhang, Zhizhong and Tan, Xin and Qu, Yanyun and Zhang, Wensheng and Han, Zhi and Xie, Yuan},
  journal={IEEE Transactions on Pattern Analysis and Machine Intelligence}, 
  year={2025},
  volume={47},
  number={10},
  pages={9316-9329},
}

@article{cheng2026toward,
  title = {Toward Sustainable On-Device Intelligence: A Survey on Energy-Efficient RAG Systems with Small Language Models},
  author = {Cheng, Zhiyuan and Lai, Longying and Liu, Yue and Sun, Yu},
  journal = {Available at SSRN 6698538},
  year = {2026}
}

@inproceedings{liu-etal-2026-ips,
  title = {{IPS}: In-Prompt Process Supervision for Short Video Content Moderation},
  author = {Liu, Mingchao and Sun, Yu and Sun, Ruixiao and Dong, Xin and Shen, Xiang and Wang, Hongwei and Xiong, Hongyu and Song, Yang},
  booktitle = {Proceedings of the 64th Annual Meeting of the Association for Computational Linguistics (Volume 6: Industry Track)},
  year = {2026},
  pages = {1277--1288},
  url = {https://aclanthology.org/2026.acl-industry.89/},
  doi = {10.18653/v1/2026.acl-industry.89},
  isbn = {979-8-89176-394-4}
}

@inproceedings{wang-etal-2025-reasoning-enhanced,
  title = {Reasoning-Enhanced Domain-Adaptive Pretraining of Multimodal Large Language Models for Short Video Content Governance},
  author = {Wang, Zixuan and Sun, Yu and Wang, Hongwei and Jing, Baoyu and Shen, Xiang and Dong, Xin and Hao, Zhuolin and Xiong, Hongyu and Song, Yang},
  booktitle = {Proceedings of the 2025 Conference on Empirical Methods in Natural Language Processing: Industry Track},
  publisher = {Association for Computational Linguistics},
  year = {2025},
  pages = {1104--1112},
  url = {https://aclanthology.org/2025.emnlp-industry.77/},
  doi = {10.18653/v1/2025.emnlp-industry.77},
  isbn = {979-8-89176-333-3}
}

@inproceedings{10.1145/3711896.3737195,
  author = {Sun, Yu and Li, Yin and Sun, Ruixiao and Liu, Chunhui and Zhou, Fangming and Jin, Ze and Wang, Linjie and Shen, Xiang and Hao, Zhuolin and Xiong, Hongyu},
  title = {Audio-Enhanced Vision-Language Modeling with Latent Space Broadening for High Quality Data Expansion},
  year = {2025},
  isbn = {9798400714542},
  publisher = {Association for Computing Machinery},
  url = {https://doi.org/10.1145/3711896.3737195},
  doi = {10.1145/3711896.3737195},
  booktitle = {Proceedings of the 31st ACM SIGKDD Conference on Knowledge Discovery and Data Mining V.2},
  pages = {4872--4881},
  numpages = {10},
  location = {Toronto, ON, Canada},
  series = {KDD '25}
}
\bibliographystyle{plainnat}

\newpage
\appendix
\section{The End-to-End Procedure}
\label{app:notation}

The notation is collected in Table~\ref{tab:notation} in the Methodology. Figure~\ref{fig:headline}(a) shows the same procedure as a diagram; Algorithm~\ref{alg:mask} states it step by step.

\section{Attribution Cost}
\label{app:attribution_cost}

\begin{algorithm}[t]
\caption{Masked continual fine-tuning with a frozen selection mask}
\label{alg:mask}
\small
\begin{algorithmic}[1]
\Require base model $f_\theta$ with frozen $W_0$; task stream $\{(\mathcal{D}_t)\}_{t=1}^{T}$; calibration set $\mathcal{D}$; budget $k$; rule $\in \{\text{gradient}, \text{uniform}\}$
\Ensure frozen mask $m$; adapted adapter $W = W_0 + \alpha BA$
\State $\mathcal{S} \gets$ the $(A,B)$ LoRA coordinates \Comment{trainable coordinates}
\If{rule $=$ gradient}
    \For{each coordinate $i \in \mathcal{S}$}
        \State $g_i \gets \mathbb{E}_{x \sim \mathcal{D}}\left[\left|\partial \mathcal{L}(x) / \partial w_i\right|\right]$ \Comment{Eq.~\eqref{eq:grad_score}}
    \EndFor
\Else
    \State $g_i \gets \mathrm{Uniform}(0,1)$ for each $i \in \mathcal{S}$
\EndIf
\State $m \gets \mathbb{1}[\,g_i \geq \tau_k\,]$ \Comment{top-$k$ rule, Eq.~\eqref{eq:topk}}
\State \textbf{freeze} $m$ \Comment{shared by all tasks; never updated}
\State $W \gets W_0$ \Comment{standard zero-init LoRA, $B = 0$}
\For{$t = 1$ to $T$}
    \State train the coordinates with $m_i = 1$ on $\mathcal{D}_t$, updating only those
    \State evaluate $\mathcal{L}_j^{(t)}$ for all $j \le t$ \Comment{every task seen so far}
\EndFor
\State \Return $m$, and $\mathrm{BWT}$ from Eq.~\eqref{eq:bwt} over $\mathcal{P} = \{1,\dots,T-1\}$
\end{algorithmic}
\end{algorithm}

\begin{table}[h]
\caption{Attribution cost on Qwen2.5-3B (BF16). Both columns are from a single \emph{full-parameter} measurement run (\texttt{--include-frozen}, \texttt{--full-precision}); the exact correspondence and its limits are described below. Peak GPU is \texttt{torch.cuda.max\_memory\_allocated} converted to decimal GB (profiling fields \texttt{*\_peak\_alloc\_mb} = 13168 / 7102 MiB, i.e. 13.8 / 7.4 GB); Time is the per-task attribution wall-clock recorded by the run (584.9\,s gradient / 12.1\,s Wanda).}
\label{tab:memory}
\centering
\small
\begin{tabular}{@{}lccc@{}}
\toprule
\textbf{Method} & \textbf{Peak GPU (GB)} & \textbf{Time (s)} & \textbf{Backward?} \\
\midrule
Gradient mask & $13.8^{\dag}$ & 584.9 & Yes \\
Wanda proxy & 7.4 & 12.1 & No \\
Shared random / RRM & \textbf{0}\textsuperscript{\ddag} & \textbf{0} & No \\
No mask & 0\textsuperscript{\ddag} & 0 & No \\
\bottomrule
\end{tabular}

\vspace{2pt}
{\footnotesize $^{\dag}$\emph{Full-parameter} attribution configuration (\texttt{--include-frozen}, scoring all 3{,}085{,}938{,}688 base parameters); no method in this paper uses this configuration. The attribution used in the continual-learning experiments records a $\sim$0.73\,GB transient allocation above the resident base model during its separate calibration pass (4-bit NF4, seq 512, 64 calibration samples, gradient checkpointing enabled; \texttt{overhead\_mb} = 699.43\,MiB). This is not an increase over the ordinary training backward peak; the removable costs are the additional calibration pass, its wall-clock compute, and its data requirement. \\ $^{\ddag}$Zero by construction because no separate attribution pass is run, not a measured end-to-end peak-memory saving.}
\end{table}

Both measured columns of Table~\ref{tab:memory} come from a single full-parameter attribution profiling run (\texttt{--include-frozen}, \texttt{--full-precision}), so they are internally consistent; that run scores all 3{,}085{,}938{,}688 base-model parameters, a configuration no reported method uses. We describe the memory figure first, then the time figure.

Gradient attribution requires 13.8\,GB peak GPU memory for Qwen2.5-3B (\texttt{max\_memory\_allocated}, 13168\,MiB $\to$ 13.8\,GB in decimal units). This figure is from the full-parameter attribution configuration, which scores all 3{,}085{,}938{,}688 parameters of the base model. The peak comprises four components (the resident base-model weights, the forward and backward activation graph over the full model, the full-parameter gradient accumulator, and the CUDA context), but these were not separately instrumented in the surviving run record, so we report the total rather than a measured decomposition. The backward pass does traverse the base model: attribution computes $\partial \mathcal{L}/\partial w$ over calibration samples, so both the forward and backward graphs span the full parameter space rather than the 29.9M-parameter LoRA surface. This configuration peaks at 13.8\,GB, of which $\sim$90\% is resident base-model weights (6.23\,GB in BF16) plus the full-model gradient buffer (6.23\,GB); these are costs the CL configuration does not pay. It is therefore an upper bound on a configuration no method uses, not a measure of attribution's own overhead. The CL attribution pass records a $\sim$0.73\,GB transient allocation above the resident base model, but ordinary training also performs backward passes; this is not an incremental end-to-end peak-memory cost. The random baseline removes the separate calibration pass and its time/data costs, not the training backward peak (Table~\ref{tab:cl_3b}).

The 584.9\,s per-task figure is the wall-clock of that same full-parameter run (\texttt{full\_alora\_time\_s} = 584.9\,s). As a checkpointing-off contrast, an earlier LoRA-only attribution profiling run (4-bit, seq 1024, 256 calibration samples, gradient checkpointing disabled) reports 3{,}512\,MiB of overhead, a different regime (checkpointing off, longer sequences) that should not be combined with the 0.73\,GB figure above.

\section{Pruning Saliency Is Orthogonal to Gradient Importance}
\label{app:pruning_saliency}

\begin{figure}[h]
\centering
\includegraphics[width=0.45\textwidth]{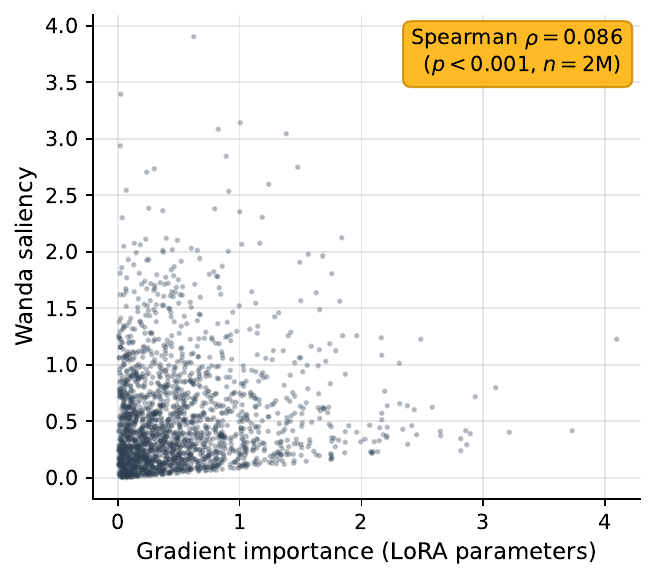}
\caption{Spearman $\rho = 0.086$ between Wanda and gradient importance (LoRA parameters) on Qwen2.5-3B. These metrics rank parameters nearly independently.}
\label{fig:spearman}
\end{figure}

Spearman correlation between Wanda ($|W| \times \text{RMS}(X)$) and gradient importance ($|\partial \mathcal{L}/\partial w|$) across 2M sampled parameters: (Figure~\ref{fig:spearman}; $\rho = 0.086$, 95\% CI $\approx [0.084, 0.088]$, $n{=}2$M; correlation-only, measured outside a continual-learning context).

\section{Confounded Null-Result Statistics}
\label{app:null_stats}

\begin{table}[h]
\caption{Per-seed BWT breakdown (Qwen2.5-3B, rank-16 LoRA, 5 tasks).}
\label{tab:per_seed}
\centering
\small
\begin{tabular}{@{}lccc@{}}
\toprule
\textbf{Seed} & \textbf{Gradient BWT $\downarrow$} & \textbf{Random BWT $\downarrow$} & \textbf{No Mask BWT $\downarrow$} \\
\midrule
42  & 0.491 & 0.435 & 0.507 \\
123 & 0.463 & 0.417 & $\mathbf{0.409}$ \\
456 & $\mathbf{0.388}$ & 0.414 & 0.539 \\
789 & $\mathbf{0.357}$ & 0.417 & 0.484 \\
1024 & 0.486 & 0.458 & 0.479 \\
\midrule
Mean & 0.437 & $\mathbf{0.428}$ & 0.484 \\
Std  & 0.061 & 0.019 & 0.047 \\
\bottomrule
\end{tabular}
\end{table}

\begin{table}[h]
\caption{Confounded comparison (Qwen2.5-3B, rank-16 LoRA, 5 tasks, $n{=}5$ seeds). The last column records whether an additional calibration forward--backward pass is required; its time and data costs are described in Appendix~\ref{app:attribution_cost}. Across all tables, bold marks the better value in its column.}
\label{tab:cl_3b}
\centering
\small
\begin{tabular}{@{}lccc@{}}
\toprule
\textbf{Method} & \textbf{BWT $\downarrow$} & \textbf{Loss $\downarrow$} & \textbf{Calibration pass?} \\
\midrule
Shared random & $\mathbf{0.428} \pm 0.019$ & $\mathbf{1.940} \pm 0.021$ & No \\
Gradient mask & $0.437 \pm 0.061$ & $1.966 \pm 0.037$ & Yes \\
No mask & $0.484 \pm 0.047$ & $1.970 \pm 0.033$ & No \\
\bottomrule
\end{tabular}
\end{table}

A paired $t$-test on per-seed BWT (Table~\ref{tab:per_seed}; random vs.\ gradient: seeds 42/123/456/789/1024 yield differences $+0.056$, $+0.046$, $-0.026$, $-0.059$, $+0.028$) gives $t(4) = 0.41$, $p = 0.70$ (two-tailed), Cohen's $d_z = 0.18$ (small effect). 
The direction is seed-dependent, consistent with no systematic advantage for either method. Normalizing BWT per task by zero-shot loss ($\Delta L / L_0$) does not change the method ranking. Tables~\ref{tab:cl_3b} and~\ref{tab:within_b} report different \texttt{full\_alora} BWT for the same seed because they come from independent runs; the absolute BWT level varies across sessions (\S\ref{app:nf4_drift}).



\section{Within-\texttt{lora\_B} Comparison (Density-Unmatched)}
\label{app:within_b}

The within-\texttt{lora\_B} comparison matches the parameter pool (both arms draw from \texttt{lora\_B}) but \emph{not} the density: the gradient mask activates approximately $2\times$ the B parameters of random-B-only. It is therefore an intermediate, density-unmatched step; the density-matched control (D1) is the headline comparison in the main text. All comparisons are within-run (same method ordering and RNG state); $\Delta = \text{random\_B\_only} - \text{Grad-Full}$, negative $\Delta$ favors random-B. Absolute BWT values carry session-level variation and should not be compared across seeds (\S\ref{app:nf4_drift}).

\begin{table}[t]
\caption{Within-\texttt{lora\_B} comparison (Qwen2.5-3B, rank-16 LoRA, 5 tasks, 7 seeds). All comparisons are within-run (same method ordering and RNG state). $\Delta = \text{random\_B\_only} - \text{Grad-Full}$; negative $\Delta$ indicates random-B achieves lower (better) BWT.}
\label{tab:within_b}
\centering
\small
\begin{tabular}{@{}lccr@{}}
\toprule
\textbf{Seed} & \textbf{random\_B\_only BWT $\downarrow$} & \textbf{full\_alora BWT $\downarrow$} & \textbf{$\Delta$} \\
\midrule
42  & $\mathbf{0.326}$ & 0.491 & $-0.165$ \\
123 & $\mathbf{0.320}$ & 0.413 & $-0.093$ \\
456 & $\mathbf{0.337}$ & 0.356 & $-0.019$ \\
789 & $\mathbf{0.346}$ & 0.364 & $-0.018$ \\
1024 & $\mathbf{0.336}$ & 0.442 & $-0.106$ \\
2048 & $\mathbf{0.367}$ & 0.402 & $-0.035$ \\
4096 & $\mathbf{0.331}$ & 0.398 & $-0.067$ \\
\midrule
Mean & $\mathbf{0.338} \pm 0.015$ & $0.409 \pm 0.046$ & $-0.072$ \\
\bottomrule
\end{tabular}
\end{table}

\section{Variation Across Runs and Method-Order Variation}
\label{app:nf4_drift}

Three quantities in this paper are sometimes read as competing estimates of one
noise level. They are not: each measures a different thing at a different
granularity, and the table below separates them.

\begin{table}[h]
\caption{Distinct quantities that are sometimes all called run-to-run \emph{noise}. The paired tests use the across-seed SD of the within-session contrasts; the protocol-equivalence discrepancy is an implementation check and does not enter inference. Position and cross-session effects are reported as sensitivities rather than as interchangeable error estimates.}
\label{tab:noise_levels}
\centering
\small
\begin{tabular}{@{}p{0.17\textwidth}p{0.17\textwidth}p{0.16\textwidth}p{0.38\textwidth}@{}}
\toprule
\textbf{Level} & \textbf{What varies} & \textbf{Magnitude} & \textbf{What it measures} \\
\midrule
(a) protocol-equivalence check & execution path only & $9.6 \times 10^{-5}$ & numerical agreement between two code paths; not an inferential error term \\
(b) across-seed paired contrast & independent seed & SD$(\Delta_s)=0.071$ & variability used by the paired significance test \\
(c) position (ordering) sensitivity & position of the two arms & $0.064$ (seed 42, \texttt{gradient\_B}) & method-position interaction in a three-seed check \\
(d) cross-session variation & session, hardware, and method & up to $0.120$ (\texttt{full\_alora}, seed 42) & method-specific offset of unknown origin \\
\bottomrule
\end{tabular}
\end{table}

\begin{table}[h]
\caption{Method-ordering sensitivity (D1 density-matched, three seeds). $\Delta_{\text{pos}}$ is the BWT difference between running at position~1 and position~2. The last two columns express each arm's positional shift as a fraction of the main D1 contrast ($|\Delta| = 0.155$).}
\label{tab:method_order}
\centering
\small
\begin{tabular}{@{}lccrr@{}}
\toprule
\textbf{Seed} & $|\Delta_{\text{pos}}(\text{rB})|$ & $|\Delta_{\text{pos}}(\text{gB})|$ & \textbf{rB / main} & \textbf{gB / main} \\
\midrule
42   & $1.5 \times 10^{-3}$ & $6.4 \times 10^{-2}$ & 0.01 & 0.41 \\
789  & $3.1 \times 10^{-5}$ & $6.2 \times 10^{-3}$ & 0.00 & 0.04 \\
2048 & $8.4 \times 10^{-6}$ & $1.3 \times 10^{-2}$ & 0.00 & 0.08 \\
\bottomrule
\end{tabular}
\end{table}

\paragraph{(a) Same process, protocol split.} Running each method in its own
process reproduces the multi-method BWT to within $9.6\times10^{-5}$, an order of
magnitude below the pre-specified $10^{-3}$ threshold. This is a check on the
protocol implementation, not on run-to-run variation of the model: two code
paths that should agree do agree. The paired $t$ tests instead use the
across-seed standard deviation of the within-session paired contrasts,
SD$(\Delta_s)=0.071$ for D1.

\paragraph{(b) Same session, arm order swapped.} Within one D1 session the two
arms run back to back, and reversing their order is a genuine perturbation of
the run. At seed 42 the \texttt{gradient\_B} arm's BWT moves by 0.064 between the
two orders; at seeds 789 and 2048 both arms
stay within $1.4\times10^{-2}$. This is a within-session positional effect, and
the within-run contrast survives it at all three tested seeds. Because this
sensitivity analysis covers three rather than seven seeds, it does not establish
the population distribution of the position effect.

\paragraph{(c) Cross-session, same seed.} The absolute BWT of a fixed method and
seed is not reproducible across sessions. For seed 42, \texttt{full\_alora}
records $0.491$, $0.378$, and $0.498$ in three separate runs (spread $0.120$),
while \texttt{random\_B\_only} records $0.326$, $0.324$, and $0.278$ (spread
$0.048$). The spread is method-dependent, so it does not act as a constant offset
that cancels in a comparison.

Method position can shift the absolute BWT of Gradient-B, most notably at seed 42, but reversing the execution order does not reverse the Random-B–Grad-B contrast in any of the three tested seeds. (Table~\ref{tab:method_order})

\begin{table}[h]
\caption{Absolute loss-based BWT for the same seed 42 under identical method
definitions, across sessions and hardware. Run C is the D1 density-matched
session and Run D the uniform-B suppression control (the M2 controller) session, both on a separate GPU and software
stack; \texttt{random\_A+B} was not run in either. A Run-D entry exists only for
methods Run D ran, and it did not train a plain \texttt{full\_alora} arm, so that
cell is \texttt{---} and its gradient-side arm is \texttt{gradient\_B}, entered on
its own row. Spread is $\max-\min$ across the runs in which the method appears.}
\label{tab:nf4_drift}
\centering
\small
\begin{tabular}{@{}lcccc@{}}
\toprule
\textbf{Method} & \textbf{Run A} & \textbf{Run B} & \textbf{Run C} & \textbf{Run D / spread} \\
\midrule
Gradient mask (\texttt{full\_alora}) & 0.491 & 0.378 & 0.498 & --- (0.120) \\
Gradient-B@10\% (\texttt{gradient\_B}) & --- & --- & 0.453 & 0.423 (0.030) \\
Random mask (\texttt{random\_A+B}) & 0.435 & 0.349 & --- & --- \\
Random B-only (\texttt{random\_B\_only}) & 0.326 & 0.324 & 0.278 & 0.278 (0.048) \\
\bottomrule
\end{tabular}
\end{table}

\paragraph{Why this is not called NF4 non-determinism.} Quantization is the same
in every run of Table~\ref{tab:nf4_drift}, so it cannot by itself explain a
cross-session spread while leaving a same-process protocol split at
$9.6\times10^{-5}$. The cross-session spread is method-specific rather than a
common shift shared by all arms. We therefore report it as a \emph{method-specific
cross-session offset of unknown origin} and do not
attribute it to a mechanism. Two consequences follow. First, only within-run
comparisons are used for any claim in this paper. Second, the \texttt{random\_B\_only}
value that enters Table~\ref{tab:d1} is the D1 session's own $0.278$, taken from the
same session as its partner arm, not an average across sessions.

\paragraph{Attribution of the level (c) spread.} The level (c) spread has no single
cause. Method position is not such a cause within a session:
\texttt{set\_seed} is called before each method's model load and no random numbers
are consumed between iterations, so the seed state entering training does not
depend on position. Level (b) shows that position can still matter, and level
(c) shows that session identity can. Which of these dominates across a full
re-run is not tested here.

Therefore, absolute BWT levels should not be compared across sessions; all statistical claims in the paper use within-session paired contrasts.

\section{RRM Boundary Analysis}
\label{app:rrm}

We investigated whether assigning each task a different random mask (Rotating Random Masks, RRM) provides additional benefit beyond shared random masking. RRM generates a deterministic per-task mask via $m^{(t)}_i = \mathbf{1}[\text{hash}(t \| i) \bmod 10 < k]$, reducing expected pairwise inter-task parameter overlap from 100\% (shared mask) to $k^2$ (1\% at 10\% sparsity) (Table~\ref{tab:rrm_boundary} ). 

\begin{table}[h]
\caption{RRM boundary analysis (Qwen2.5-3B, single seed per condition except rank-16 reference).}
\label{tab:rrm_boundary}
\centering
\small
\begin{tabular}{@{}lccrl@{}}
\toprule
\textbf{Condition} & \textbf{RRM BWT} & \textbf{Random BWT} & \textbf{$\Delta$} & \textbf{Direction} \\
\midrule
Rank 4 & 0.464 & 0.423 & $+$0.041 & Random better \\
Rank 8 & 0.427 & 0.373 & $+$0.055 & Random better \\
Rank 16 (ref.) & 0.401 & 0.435 & $-$0.034 & RRM better \\
Rank 32 & 0.414 & 0.390 & $+$0.024 & Random better \\
Reverse order & 0.115 & 0.109 & $+$0.006 & Random better \\
\bottomrule
\end{tabular}
\end{table}

RRM shows a favorable paired difference at rank 16 only. The overlap-reduction hypothesis predicts a consistent RRM advantage; the data do not support this across conditions. RRM's effect appears condition-dependent rather than robust.

\section{Sparsity Sweep}
\label{app:sparsity}

We sweep the masking budget over 5\%, 10\%, 20\%, and 30\% (single seed) to test whether the within-pool ordering is specific to the 10\% sparsity used in the main experiments. Across all four levels, shared random achieves numerically lower BWT than gradient attribution, with no cross-over; the largest margin is at 30\%.

\begin{figure}[ht]
\centering
\includegraphics[width=0.55\textwidth]{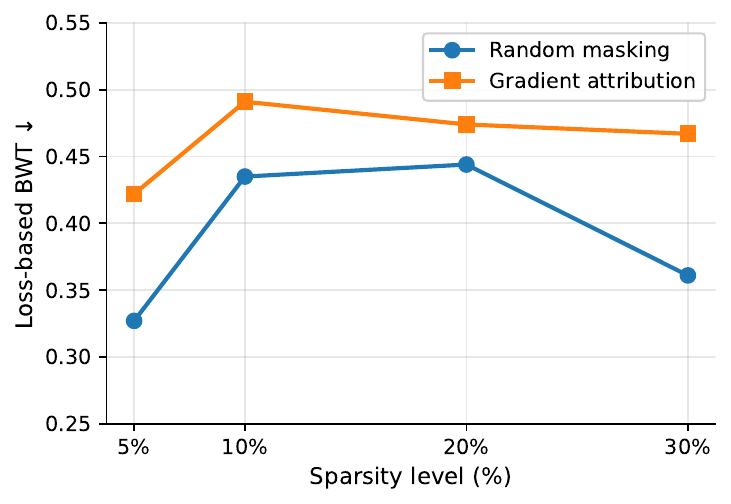}
\caption{Sparsity sweep (Qwen2.5-3B, single seed). Shared random achieves numerically lower BWT than gradient attribution at all four tested levels.}
\label{fig:sparsity}
\end{figure}

\begin{table}[h]
\caption{Sparsity sweep numerical values (Qwen2.5-3B, single seed). Random BWT $\leq$ Gradient BWT at all tested levels.}
\label{tab:sparsity}
\centering
\small
\begin{tabular}{@{}lcc@{}}
\toprule
\textbf{Sparsity} & \textbf{Random BWT $\downarrow$} & \textbf{Gradient BWT $\downarrow$} \\
\midrule
5\% & 0.327 & 0.422 \\
10\% & 0.435 & 0.491 \\
20\% & 0.444 & 0.474 \\
30\% & 0.361 & 0.467 \\
\bottomrule
\end{tabular}
\end{table}

\section{Training Stability: Rank-8 No-Mask Divergence}
\label{app:training_stability}

The rank-8 no-mask ablation reveals a discrepancy between two endpoint-based forgetting summaries. Endpoint BWT is 0.176; average forgetting is 0.2837. Training produced a severe loss spike during task 4 (Math): evaluation losses on three prior tasks exceeded 14, compared to values below 4 for all masked conditions at the same rank. This spike subsequently recovered during task 5, yielding deceptively low endpoint BWT. We do not interpret the rank-8 no-mask endpoint BWT of 0.176 as evidence that no-mask outperforms masking.

The same signature recurs at rank 16 and is seed-dependent: at seed 123 the unmasked arm spiked during task 4 with evaluation losses of 16.2/16.0/14.7 on the three prior tasks (code/medical/legal), while every masked arm in that run stayed below $2.5$; at seed 42 the unmasked arm did \emph{not} spike (task-4 evaluations $\leq 1.2$). The instability is therefore a property of the unmasked (all-parameter) configuration rather than of a particular rank, and it appears in a minority of seeds.

\section{Mask Concentration by Module}
\label{app:mask_concentration}

The o\_proj suppression ablation and module-concentration figures, discussed in \S\ref{sec:analysis}, are reproduced below together with the module-concentration table for reference.

\begin{figure}[t]
\centering
\includegraphics[width=0.6\textwidth]{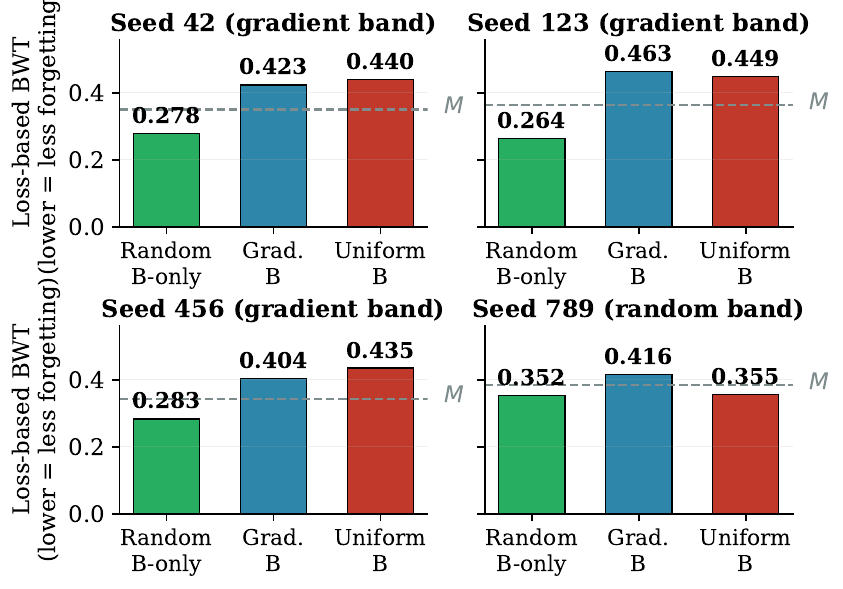}
\caption{The suppression control's outcome is \emph{seed-dependent}. All three arms train 1{,}651{,}508 parameters, which is 5.5\% of the full LoRA pool (each arm draws 10\% of the \texttt{lora\_B} sub-pool, and $B$ holds 55\% of the pool's parameters), so this is the density-matched setting throughout. The uniform-B controller redistributes the gradient mask's budget uniformly across LoRA modules. At seed 42 it sits in the gradient band (0.440 above gradient-B's 0.423 and well above random-B-only's 0.278); seeds 123 and 456 also land on the gradient side; at seed 789 it falls to the random band (0.355 below gradient-B's 0.416, alongside random-B-only's 0.352). Three of four seeds fall on the gradient side. M2 is an independent session from D1, so the same-seed gradient-B values differ between them (e.g.\ seed 42: 0.423 here vs 0.453 in the D1 table; seed 789: 0.416 vs 0.387; \S\ref{app:nf4_drift}). We therefore do not treat this ablation as establishing the \texttt{o\_proj}-concentration mechanism (\S\ref{sec:analysis}).}
\label{fig:oproj_ablation_main}
\end{figure}

\begin{figure}[t]
\centering
\includegraphics[width=0.5\textwidth]{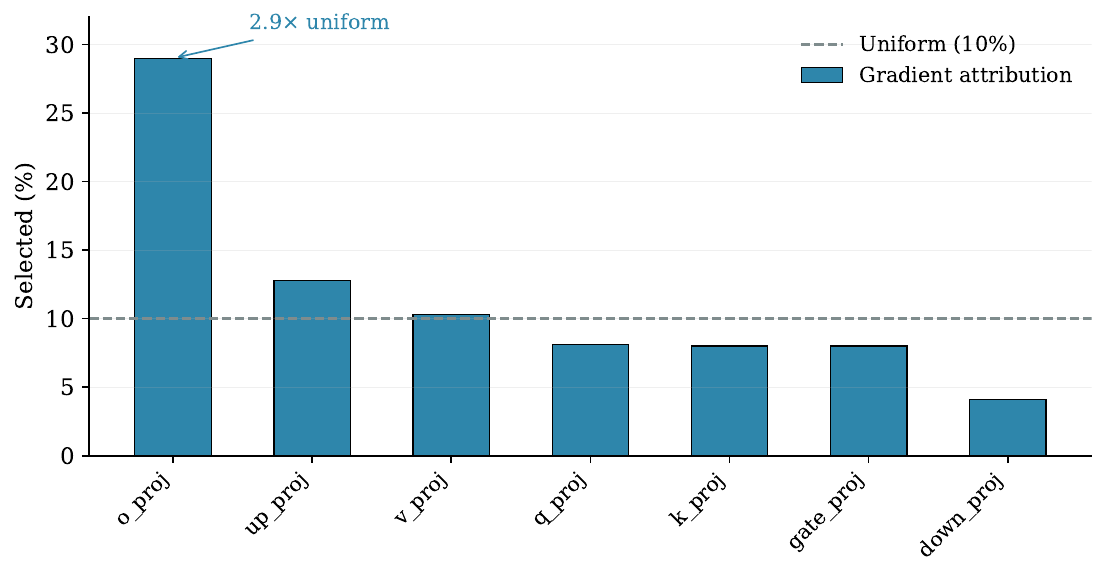}
\caption{Gradient-mask concentration by LoRA module (Qwen2.5-3B). Gradient attribution selects 29\% of \texttt{o\_proj} parameters (2.9$\times$ the 10\% uniform rate) and is under-represented in \texttt{down\_proj} (4.1\%); the dashed line marks uniform 10\% selection.}
\label{fig:module_concentration_main}
\end{figure}

\begin{table}[t]
\caption{Gradient mask concentration by LoRA module (Qwen2.5-3B). Random masking distributes uniformly at 10\% per module. Gradient attribution selects exclusively \texttt{lora\_B} parameters (all \texttt{lora\_A} receive zero importance due to $B=0$ initialization).}
\label{tab:mask_concentration}
\centering
\small
\begin{tabular}{@{}lccr@{}}
\toprule
\textbf{Module} & \textbf{Gradient density} & \textbf{Random density} & \textbf{Ratio} \\
\midrule
o\_proj & 29.0\% & 10.0\% & 2.9$\times$ \\
up\_proj & 12.8\% & 10.0\% & 1.3$\times$ \\
v\_proj & 10.3\% & 10.0\% & 1.0$\times$ \\
q\_proj & 8.1\% & 10.0\% & 0.8$\times$ \\
k\_proj & 8.0\% & 10.0\% & 0.8$\times$ \\
gate\_proj & 8.0\% & 10.0\% & 0.8$\times$ \\
down\_proj & 4.1\% & 10.0\% & 0.4$\times$ \\
\bottomrule
\end{tabular}
\end{table}

\section{Generalization and Boundary Checks}
\paragraph{Scale Observations}
\label{app:scale}

We additionally observe the within-pool comparison at three parameter scales (0.5B, 1.5B, 7B; single seed each). These are observational rather than multi-seed, and at 0.5B and 1.5B only the shared-random and no-mask conditions were run, so no gradient attribution baseline is available for a direct comparison there.

\begin{table}[ht]
\caption{BWT at additional scales (single seed, observational).}
\label{tab:scale_obs}
\centering
\small
\begin{tabular}{@{}lccc@{}}
\toprule
\textbf{Scale} & \textbf{Random BWT $\downarrow$} & \textbf{Gradient BWT $\downarrow$} & \textbf{No Mask BWT $\downarrow$} \\
\midrule
0.5B & 0.343 & --- & 0.431 \\
1.5B & 0.309 & --- & 0.356 \\
7B & 0.447 & 0.441 & 0.514 \\
\bottomrule
\end{tabular}
\end{table}

\paragraph{Task Ordering (single seed)}
\label{app:task_order}

A reversed-order experiment yields BWT $\approx 0.11$ for all masking methods, compared to $\approx 0.42$ under the standard order, a nearly $4{\times}$ reduction that far exceeds the method differences measured within this experiment (within-session method contrasts of order $10^{-2}$). This is a \emph{single seed}, so the $4{\times}$ magnitude is a qualitative pointer, not a quantitative estimate; it is reported to show that task ordering dominates the forgetting landscape, not as a measured effect size. The conclusions of this paper apply only to the tested Code$\to$Medical$\to$Legal$\to$Math$\to$Creative sequence; additional task orderings with multi-seed replication are needed before the null result can be generalized.

\paragraph{Cross-Architecture Comparison}
\label{app:cross_arch}

Both cross-architecture designs are tabulated in Table~\ref{tab:cross_arch}: the confounded design (gradient selecting B-only, random selecting A+B) and the within-\texttt{lora\_B} design (both methods restricted to \texttt{lora\_B}). The two designs are not comparable row-for-row because the Phi-2 within-B rows use a different LoRA target set; \S\ref{sec:boundaries} discusses both.

\begin{table}[t]
\caption{Cross-architecture comparison on Qwen2.5-3B and Phi-2 (2.78B, GPT-NeoX-style), with both arms within a row sharing one target set. Qwen2.5-3B targets \texttt{q/k/v/o/gate/up/down\_proj} (pool $29{,}933{,}568$), whereas Phi-2 uses GPT-NeoX-style \texttt{q,k,v,fc1,fc2,dense} projections (pool $23{,}592{,}960$) in the confounded design. In the within-\texttt{lora\_B} design, the Phi-2 rows omit \texttt{dense} (pool $20{,}971{,}520$, 12.5\% smaller), so the two Phi-2 designs are not comparable row-for-row. The within-\texttt{lora\_B} rows are matched on pool but \emph{not} density (the gradient mask activates approximately $2\times$ the B parameters of Random-B), as on Qwen2.5-3B.}
\label{tab:cross_arch}
\centering
\small
\begin{tabular}{@{}llccc@{}}
\toprule
\textbf{Architecture} & \textbf{Comparison} & \textbf{Grad.\ BWT $\downarrow$} & \textbf{Rand.\ BWT $\downarrow$} & \textbf{$\Delta$} \\
\midrule
Qwen2.5-3B ($n{=}5$) & Confounded & $0.437 \pm 0.061$ & $0.428 \pm 0.019$ & $-0.009$ \\
Phi-2 ($n{=}5$) & Confounded & $0.158 \pm 0.019$ & $0.183 \pm 0.016$ & $+0.025$ \\
\midrule
Qwen2.5-3B ($n{=}7$) & Within-B & $0.409 \pm 0.046$ & $0.338 \pm 0.015$ & $-0.072$ \\
Phi-2 ($n{=}4$) & Within-B & $0.149 \pm 0.036$ & $0.065 \pm 0.017$ & $-0.084$ \\
\bottomrule
\end{tabular}
\end{table} 

\section{In-Domain Calibration Ablation}
\label{app:calibration}

The primary experiments use out-of-domain Alpaca calibration data (64 samples). To test whether this choice drives the null result, we repeat the gradient-mask experiment with \emph{pooled in-domain} calibration: 60 samples drawn equally from the five CL tasks.\footnote{$\lfloor 64/5 \rfloor \times 5 = 60$; a 6.25\% size difference from integer division.} All other parameters are identical (Qwen2.5-3B, seed 42, rank 16, 10\% sparsity).

Loss-based BWT is 0.495 vs 0.491 with Alpaca calibration ($\Delta = 0.004$, 0.8\% relative), well within the 5-seed standard deviation of 0.061. Random and no-mask controls are numerically identical across both runs, confirming infrastructure determinism. The \emph{cross-calibration} overlap (Alpaca vs pooled) is Jaccard $= 0.585$: changing the calibration domain alters 42\% of selected parameters. Despite this substantial mask difference, BWT changes by only 0.004. This robustness is of the \emph{outcome}, not the mask: the mask does change, but \emph{among gradient-derived masks} the specific 10\% selected has little influence on forgetting, whereas swapping the selection criterion entirely (gradient for random) does, as the density-matched control shows (\S\ref{sec:confounded}).

\section{B=0 Restriction Asymmetry}
\label{app:b0_asymmetry}

The within-\texttt{lora\_B} comparison varies the selection criterion while holding the parameter pool fixed. That the pool restriction is the right control to hold follows from an asymmetry between the two arms at seed 42 (Table~\ref{tab:b0_asymmetry}). Restricting the gradient mask to \texttt{lora\_B} changes nothing, because under $B=0$ every \texttt{lora\_A} importance score is already zero and the restriction is a no-op. The same restriction applied to the random arm removes the \texttt{lora\_A} dilution and improves BWT by 0.025. A restriction that does nothing to one arm and helps the other is evidence that the two arms were drawing from different pools before the restriction was applied, which is what makes the pool-matched comparison the informative one.

\begin{table}[h]
\centering\small
\caption{Effect of restricting the mask to \texttt{lora\_B} (seed 42). For gradient attribution, the restriction is a no-op by construction (B=0 collapses every \texttt{lora\_A} importance score to zero); for random, restricting to B removes \texttt{lora\_A} dilution and improves BWT by 0.025. This asymmetry demonstrates that the within-B comparison varies the selection criterion, not the parameter pool.}
\label{tab:b0_asymmetry}
\begin{tabular}{@{}lccc@{}}
\toprule
\textbf{Method} & \textbf{Unrestricted BWT} & \textbf{B-only BWT} & \textbf{Gap} \\
\midrule
Gradient & 0.378 & 0.378 & $+0.000$ \\
Random   & 0.349 & 0.324 & $-0.025$ \\
\bottomrule
\end{tabular}
\end{table}

\section{Per-Seed Accuracy and Matched Loss Contrast}
\label{app:accuracy_seeds}

The accuracy evaluation is reported in the main text, \S\ref{sec:accuracy_branch}, together with the option-scoring protocol; its per-seed values and the matched loss contrast are tabulated below. (Table~\ref{tab:accuracy_branch})

The interval is wide because the per-seed differences are wide, not because the mean is large. With the observed across-seed standard deviation of 4.17 pp, the 95\% interval half-width is about 3.9 pp at $n=7$ and would fall below 2.0 pp only at roughly twenty seeds. 

\begin{table}[t]
\caption{Per-seed results, both columns in the direction [Grad $-$ Random] (positive = gradient forgets more). Col.~2 is the accuracy-BWT difference: the difference in mean final-minus-learned accuracy over the prior discrete-option tasks, not a difference of final accuracies. It is therefore a retention quantity on the same footing as the loss column and does not mix learning with retention. Col.~2 is reported in percentage points and is measured in the A800 accuracy runs; Col.~3 is the loss-based BWT difference from the A800 loss runs. Together they form a second environment relative to the original V100 loss runs. Col.~2 carries the 95\% $t$-interval on the seven-seed mean. Col.~3 has mean $+0.152$ (SD $0.0749$, $t=5.37$, CI $[+0.083,+0.221]$, 7/7 same direction). The V100 loss contrast is $+0.155$ under the same [Grad $-$ Random] convention. Across matched seeds, the A800-minus-V100 loss-contrast difference has mean $-0.0033$ and SD $0.0441$, so individual seeds move between environments even though the average does not.}
\label{tab:accuracy_branch}
\centering
\small
\begin{tabular}{@{}lrr@{}}
\toprule
\textbf{Seed} & \textbf{Accuracy-BWT $\Delta$ [Grad $-$ Rand], pp} & \textbf{Loss BWT $\Delta$ [Grad $-$ Rand]} \\
\midrule
42   & $+2.00$ & $+0.2320$ \\
123  & $0.00$  & $+0.1647$ \\
456  & $+0.48$ & $+0.1300$ \\
789  & $-6.98$ & $+0.0786$ \\
1024 & $+4.60$ & $+0.0514$ \\
2048 & $+0.51$ & $+0.2572$ \\
4096 & $+6.00$ & $+0.1492$ \\
\midrule
\multicolumn{3}{@{}l}{\emph{Summary (column 2 in percentage points, column 3 in BWT units)}} \\
Mean & $+0.94$\,pp & $+0.152$ \\
SD   & $4.17$\,pp & $0.0749$ \\
95\% CI & $[-2.91, +4.80]$\,pp & $[+0.083, +0.221]$ \\
\bottomrule
\end{tabular}
\end{table}

\section{The Math Task: Truncation Audit}
\label{app:math_audit}

Loss on math decreases with training while accuracy on the same task collapses to zero for all arms: peak accuracy at math's own training phase is $0.04$ and every arm ends at $0.00$. The numeric-match function accepts a bare number, a number with surrounding text, and a complete solution, and the prompt prefixes are identical; nevertheless, the interaction between output format and the 256-token limit makes the generation-based score unreliable. A generation audit reports 56--95\% of generations across all tasks and arms reaching the cap without an end-of-sequence token, and for math specifically 81--90\% of generations reach the cap. The last number in a truncated solution is often an intermediate quantity, so the reference answer is absent from the value the scorer inspects. Every arm shares the same cap and scorer, so the bias is common rather than arm-specific. We do not treat the math column as a forgetting result: under option-based scoring math yields no score, and under generation scoring it is confounded with truncation.

\section{Positioning Against Gradient-Attribution Families}
\label{app:positioning}

Table~\ref{tab:positioning} places our contribution against the gradient-attribution families we are positioned against, on the two axes that drive the study.

\begin{table}[h]
\centering
\small
\caption{Positioning against gradient-attribution families. ``B=0 pool'' marks a method whose attribution is computed at a zero-initialized LoRA, where the gradient with respect to \texttt{lora\_A} is zero and the mask is confined to \texttt{lora\_B}; ``within-pool grad-vs-random'' marks a method compared against a matched random baseline drawn from the same pool. Attribution-Guided CL computes its attribution from a separately trained model ($B\neq0$ at scoring time) and gates gradients continuously rather than selecting a binary mask, so it is outside the first axis by construction and unmatched on the second.}
\label{tab:positioning}
\begin{tabular}{@{}lcc@{}}
\toprule
\textbf{Method} & \textbf{B=0 pool} & \textbf{Within-pool grad-vs-random} \\
\midrule
EWC / SI / MAS & No & No \\
PiSSA ($B\neq0$) & N/A by init & No \\
Attribution-Guided CL (post-training attribution) & N/A by design & No \\
\textbf{Ours (within-\texttt{lora\_B})} & \textbf{Yes} & \textbf{Yes} \\
\bottomrule
\end{tabular}
\end{table}

Table~\ref{tab:rule_positioning} lists the selection rules this paper is positioned against and marks which we train under one protocol. Only the rules we train contribute empirical numbers here; the rest are compared by their published design, with none of their reported numbers presented as a measurement of ours.

\begin{table}[h]
\centering
\footnotesize
\caption{Selection rules this paper is positioned against. The last column marks the four rules we train and evaluate under one protocol (\S\ref{sec:mask}); the rest are compared by design only, and none of their published numbers is cited as a result of ours.}
\label{tab:rule_positioning}
\begin{tabular}{@{}p{0.21\textwidth}p{0.59\textwidth}c@{}}
\toprule
\textbf{Rule} & \textbf{What it scores, and what for} & \textbf{Run} \\
\midrule
EWC~\citep{kirkpatrick2017overcoming}; MAS~\citep{aljundi2018memory} & Fisher information, output sensitivity; a penalty, not a mask & no \\
SNIP~\citep{lee2019snip}; GraSP~\citep{wang2020grasp} & gradient at initialisation; one-shot pruning mask & no \\
Wanda~\citep{sun2024wanda} & weight$\times$activation magnitude; one-shot pruning mask & no \\
LoRA-FA~\citep{zhang2023lorafa} & none, structural; freezes \texttt{lora\_A}, trains \texttt{lora\_B} & no \\
Grad-Full, Grad-B (\S\ref{sec:mask}) & calibration gradient magnitude; static CL mask, full pool (Grad-Full) or \texttt{lora\_B} at 10\% (Grad-B) & \textbf{yes} \\
Random-B; Random A+B~\citep{su2024random} & none, uniform; static CL mask, \texttt{lora\_B} at 10\% (Random-B) or full pool (A+B) & \textbf{yes} \\
\bottomrule
\end{tabular}
\end{table}

\section{D1 Per-Task and Per-Seed Decomposition}
\label{app:d1_per_task}

Table~\ref{tab:d1_per_task_seeds} gives the same quantity for each
seed, so the task-level pattern can be checked against the seed-level one; the
main text reports the same decomposition as a figure (Figure~\ref{fig:per_task}). Each
entry is $\text{loss\_matrix}[4][j] - \text{loss\_matrix}[j][j]$ for prior task
$j$; the final column is the mean over the four tasks, which is the per-seed
aggregate $\Delta$ of Table~\ref{tab:d1}.


\begin{table}[h]
\caption{Per-seed, per-task D1 decomposition ($\Delta = \text{random-B} - \text{gradient-B}$ for each prior task). Negative favours random-B. Legal is the only task whose sign is not consistent across seeds.}
\label{tab:d1_per_task_seeds}
\centering
\small
\begin{tabular}{@{}lrrrrr@{}}
\toprule
\textbf{Seed} & \textbf{Code} & \textbf{Medical} & \textbf{Legal} & \textbf{Math} & \textbf{Aggregate} \\
\midrule
42   & $-0.040$ & $-0.409$ & $-0.002$ & $-0.249$ & $-0.175$ \\
123  & $-0.002$ & $-0.698$ & $-0.002$ & $-0.052$ & $-0.188$ \\
456  & $-0.074$ & $-0.381$ & $-0.004$ & $-0.033$ & $-0.123$ \\
789  & $-0.063$ & $-0.054$ & $+0.001$ & $-0.023$ & $-0.035$ \\
1024 & $-0.044$ & $-0.410$ & $-0.011$ & $-0.021$ & $-0.121$ \\
2048 & $-0.041$ & $-0.950$ & $+0.010$ & $-0.062$ & $-0.261$ \\
4096 & $-0.059$ & $-0.568$ & $-0.014$ & $-0.091$ & $-0.183$ \\
\midrule
Mean & $-0.046$ & $-0.496$ & $-0.003$ & $-0.076$ & $-0.155$ \\
\bottomrule
\end{tabular}
\end{table}

\section{Ruling Out a Learning Deficit in the D1 Gap}
\label{app:d1_diagonal}

The D1 density-matched comparison (both arms 10\% of \texttt{lora\_B}, 1{,}651{,}508 parameters) shows random-B-only achieving lower BWT than gradient-B. One rival reading is that gradient-B's top-10\% is the more concentrated half of the original 20\% mask, so gradient-B learns each task less well and the gap is a learning deficit rather than a retention difference. The per-task diagonal losses (each task's loss immediately after it is learned) exclude this: gradient-B's diagonal loss is lower than random-B-only's in \emph{all} 35 seed$\times$task cells (7 seeds $\times$ 5 tasks), with no exceptions.

\begin{table}[ht]
\caption{Diagonal loss (mean $\pm$ std over the seven D1 seeds), by task. Lower is better. gradient-B is lower than random-B-only in every cell (Table~\ref{tab:d1_diagonal}); Grad-Full is the uncontrolled reference. Loss diagonals only; the D1 run records contain no accuracy matrix (accuracy is measured separately in the accuracy branch, \S\ref{sec:accuracy_branch}).}
\label{tab:d1_diagonal}
\centering
\small
\begin{tabular}{@{}lccc@{}}
\toprule
\textbf{Task} & \textbf{random\_B\_only} & \textbf{gradient\_B} & \textbf{full\_alora} \\
\midrule
code     & $0.952 \pm 0.111$ & $\mathbf{0.901 \pm 0.100}$ & $0.898 \pm 0.099$ \\
medical  & $1.190 \pm 0.040$ & $\mathbf{1.152 \pm 0.040}$ & $1.144 \pm 0.035$ \\
legal    & $0.388 \pm 0.035$ & $\mathbf{0.368 \pm 0.038}$ & $0.365 \pm 0.037$ \\
math     & $3.838 \pm 0.103$ & $\mathbf{3.652 \pm 0.096}$ & $3.642 \pm 0.100$ \\
creative & $2.104 \pm 0.053$ & $\mathbf{2.067 \pm 0.053}$ & $2.056 \pm 0.054$ \\
\bottomrule
\end{tabular}
\end{table}

A second rival reading is that BWT, being a difference of losses, is inflated for gradient-B simply because it starts from a lower per-task loss. The endpoint closes this from both sides. Because BWT is the mean over prior tasks of (final loss $-$ learned loss) and the start is the mean learned loss over those same tasks, the endpoint is exactly the mean final held-out loss over the first four tasks, read directly from the final row of the stored loss matrix rather than reconstructed. gradient-B starts \emph{lower} yet ends \emph{higher} in 6 of 7 seeds, so its larger BWT reflects greater forgetting rather than a lower baseline (Table~\ref{tab:d1_endpoint}). Measured on the same seven within-session pairs, the endpoint difference averages $+0.0813$ (SD $0.0710$; paired $t(6)=3.03$, $p=0.023$, 95\% CI $[+0.016,+0.147]$, Cohen's $d_z=1.15$). The second environment agrees in both mean and direction: $+0.078$ ($t(6)=2.62$, $p=0.039$, 6/7), a mean separation of $0.003$ from the V100 estimate. Thus Random-B's advantage is not confined to a BWT normalization: it also achieves lower final held-out loss on the prior tasks.

The endpoint significance is less robust than the headline BWT contrast to single-seed deletion. Three of the seven leave-one-out fits exceed $p=0.05$: dropping seed 42 or 123 gives $p=0.058$, and dropping seed 4096 gives $p=0.059$, all with a mean of about $+0.076$. The remaining four deletions give $p\leq0.048$. We therefore treat the endpoint as a directional result supported by the paired design rather than as an effect that survives arbitrary single-seed removal; the density-matched BWT contrast on the same seven pairs is robust to every single-seed deletion ($p\leq0.005$ throughout).

\begin{table}[ht]
\caption{Mean final held-out loss over the four prior tasks, per seed, read from the final row of the stored loss matrix. Start is the same quantity at each task's own training step; end is at the end of the sequence. gradient-B's endpoint is higher in 6 of 7 seeds (seed 789 is the exception); paired $t(6)=3.03$, $p=0.023$, mean $+0.081$ on the V100, and $+0.078$ ($p=0.039$) on the A800. Three of the seven single-seed deletions fall above $p=0.05$ (seeds 42, 123, 4096); see text.}
\label{tab:d1_endpoint}
\centering
\small
\begin{tabular}{@{}lcccc@{}}
\toprule
\textbf{Seed} & \textbf{rB start} & \textbf{rB end} & \textbf{gB start} & \textbf{gB end (diff)} \\
\midrule
42   & 1.543 & 1.821 & 1.478 & 1.931 ($+0.110$) \\
123  & 1.632 & 1.896 & 1.553 & 2.006 ($+0.110$) \\
456  & 1.575 & 1.858 & 1.494 & 1.900 ($+0.042$) \\
789  & 1.548 & 1.900 & 1.475 & 1.862 ($-0.038$) \\
1024 & 1.566 & 1.918 & 1.489 & 1.962 ($+0.045$) \\
2048 & 1.665 & 1.896 & 1.587 & 2.078 ($+0.182$) \\
4096 & 1.618 & 1.913 & 1.553 & 2.031 ($+0.118$) \\
\bottomrule
\end{tabular}
\end{table}

Both checks are loss-based (accuracy is measured separately in the accuracy branch, \S\ref{sec:accuracy_branch}), use a single task order, and cover Qwen2.5-3B only. The endpoint is read directly from the final row of the stored matrices, so its per-seed differences are load-bearing rather than derived; the caveat above concerns the $n=7$ fragility of the full-set significance, not the definition of the quantity.

\section{Cross-Architecture Density-Matched Replication (Phi-2)}
\label{app:phi2_d1}

The density-matched contrast of \S\ref{sec:confounded} is also run on Phi-2 (2.78B, GPT-NeoX-style), an architecturally distinct family from Qwen2.5-3B. As in the primary D1 comparison, both arms select from \texttt{lora\_B} alone and use the same density: each activates exactly 1{,}048{,}576 coordinates, which is 10\% of the 10{,}485{,}760-coordinate \texttt{lora\_B} sub-pool (and 5.0\% of the 20{,}971{,}520 trainable $A{+}B$ pool). The candidate pool and update budget are therefore matched, leaving only the ranking rule different between Random-B and Grad-B.

\begin{table}[t]
\caption{Density-matched D1 replication on Phi-2, four seeds. Both arms select the same number of coordinates from the same \texttt{lora\_B} pool; $\Delta=\text{Random-B}-\text{Grad-B}$ in loss-based BWT, so negative values mean the random mask forgets less.}
\label{tab:phi2_d1}
\centering
\small
\begin{tabular}{@{}lccc@{}}
\toprule
\textbf{Seed} & \textbf{Random-B BWT $\downarrow$} & \textbf{Grad-B BWT $\downarrow$} & \textbf{$\Delta$} \\
\midrule
42   & $\mathbf{0.0397}$ & 0.1661 & $-0.1264$ \\
123  & $\mathbf{0.0363}$ & 0.1273 & $-0.0910$ \\
456  & $\mathbf{0.0461}$ & 0.1749 & $-0.1289$ \\
1024 & $\mathbf{0.0480}$ & 0.1624 & $-0.1144$ \\
\midrule
Mean & --- & --- & $-0.1152$ (SD $0.0173$) \\
\bottomrule
\end{tabular}
\end{table}

The direction replicates in all four seeds (Table~\ref{tab:phi2_d1}). The mean paired contrast is $-0.1152$ (SD $0.0173$), with paired $t(3)=-13.32$, $p=0.0009$, and 95\% CI $[-0.143,-0.088]$. A two-sided sign test over four concordant seeds gives $p=0.125$; because $n=4$ is too small for a strong distributional claim and the normality assumption behind the paired $t$-test is weak, we treat the $t$ statistic as a supporting parametric summary rather than resting the replication claim on it. The replication claim is instead based on the density-matched design and the 4/4 directional agreement.

\begin{table}[t]
\caption{Density-matched D1 contrast on both architectures under the same sign convention, $\Delta=\text{Random-B}-\text{Grad-B}$ (negative means Random-B forgets less). Phi-2 is an architecturally distinct four-seed replication of the seven-seed Qwen2.5-3B result.}
\label{tab:phi2_d1_compare}
\centering
\small
\begin{tabular}{@{}lccccc@{}}
\toprule
\textbf{Architecture} & \textbf{$n$} & \textbf{Mean $\Delta$} & \textbf{SD} & \textbf{$t$} & \textbf{$p$} \\
\midrule
Qwen2.5-3B & 7 & $-0.1552$ & 0.0707 & $-5.80$ & 0.0011 \\
Phi-2      & 4 & $-0.1152$ & 0.0173 & $-13.32$ & 0.0009 \\
\bottomrule
\end{tabular}
\end{table}

Table~\ref{tab:phi2_d1_compare} places the clean Phi-2 replication beside the primary Qwen2.5-3B result under the same sign convention. The task-level decomposition also resembles the primary model rather than merely sharing the aggregate sign. Averaged over the four Phi-2 seeds, Medical contributes 67.2\% of the aggregate contrast, Math 24.1\%, Code 5.8\%, and Legal 2.9\%; Medical is the largest contributor in every seed. On Qwen2.5-3B, Medical contributes 79.9\% of the matched contrast. Thus the two architectures agree on where most of the effect is concentrated, although the concentration is somewhat less extreme on Phi-2.

The replication uses the same five task domains, LoRA rank, calibration set, and 10\% \texttt{lora\_B} budget as D1. It differs in model family, in the Phi-2 LoRA target set (which omits \texttt{dense}, giving a 20{,}971{,}520-parameter trainable pool), and in the number of seeds. The earlier Phi-2 within-B comparison in Appendix~\ref{app:cross_arch} remains a density-unmatched boundary check; the results here are the clean cross-architecture counterpart to the headline density-matched D1 comparison.

\end{document}